\documentclass[letterpaper]{article} 
\usepackage{aaai2027} 
\nocopyright
\usepackage[hyphens]{url} 
\usepackage{graphicx} 
\newcommand{\extremeglyph}{\raisebox{-0.08ex}{\includegraphics[height=1.05em]{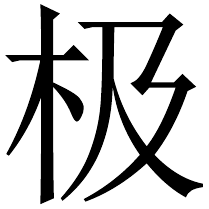}}}
\usepackage{natbib} 
\usepackage{caption} 
\usepackage{amsmath}
\usepackage{amssymb}
\usepackage{amsthm}
\usepackage{algorithm}
\usepackage{algorithmic}
\usepackage{newfloat}
\usepackage{listings}
\usepackage{booktabs}
\usepackage{subcaption}
\usepackage{tikz}
\usetikzlibrary{arrows.meta,positioning}
\DeclareCaptionStyle{ruled}{labelfont=normalfont,labelsep=colon,strut=off}
\floatstyle{ruled}
\newfloat{listing}{tb}{lst}{}
\floatname{listing}{Listing}

\theoremstyle{plain}

\theoremstyle{definition}

\theoremstyle{remark}

\title{Localizing and Correcting Errors for LLM-Based Planners}
\author{
    Aditya Kumar\textsuperscript{\rm 1}\corresponding,
    William W. Cohen\textsuperscript{\rm 1}
}
\affiliations{
    \textsuperscript{\rm 1}Machine Learning Department, Carnegie Mellon University\\
    Pittsburgh, PA 15213 USA\\
    adityaku@cs.cmu.edu, wcohen@cs.cmu.edu
}

\begin{document}

\maketitle

\begin{abstract}
Large language models (LLMs) have demonstrated strong reasoning capabilities on math and coding, but frequently fail on symbolic classical planning tasks. 
Our studies, as well as prior work, show that LLM-generated plans routinely violate domain constraints given in their instructions (e.g., walking through walls).
To address this failure, we propose iteratively augmenting instructions with Localized In-Context Learning (L-ICL) demonstrations: targeted corrections for specific failing steps. Specifically, L-ICL identifies the \emph{first} constraint violation in a trace and injects a \emph{minimal input-output example} giving the correct behavior for the failing step. Our proposed technique of L-ICL is much effective than explicit instructions or traditional ICL, which adds complete problem-solving trajectories, and many other baselines. For example, on an 8×8 gridworld, L-ICL produces valid plans 89\% of the time with only 60 training examples, compared to 59\% for the best baseline, an increase of 30 percentage points. We further find that L-ICL remains effective with imperfect correction outputs, including randomly corrupted and LLM-generated outputs, although performance declines as feedback becomes less reliable. L-ICL also shows dramatic improvements in other domains (gridworld navigation, mazes, Sokoban, and BlocksWorld), and across several LLMs.
\end{abstract}
\section{Introduction}
\label{sec:intro}

Large language models (LLMs) and agentic systems reason and plan effectively in domains such as mathematics, coding, and question answering~\cite{khattab2023dspycompilingdeclarativelanguage, yao2023tree}, suggesting that modern LLMs possess strong general planning capabilities. However, studies on classical planning benchmarks reveals a more nuanced picture: LLMs frequently fail, even on simple tasks that symbolic planners solve easily~\cite{valmeekam2023planning, stechly2024chain}.
Prior works have analyzed plans produced by LLMs such as SearchFormer \cite{lehnert2024beyond}, which are fine-tuned to generate structured reasoning chains that can be parsed, and shown that LLMs frequently \emph{violate domain constraints} given in their instructions \cite{stechly2024chain}.  For example, LLMs might propose plans that walk through a wall in a maze, or pick up a block when the robot's gripper is already occupied. 
\begin{table}[t]
\centering
{
\small
\setlength{\tabcolsep}{1mm}
\begin{tabular}{@{}lccc@{}}
\toprule
\textbf{Method} & \textbf{\% Valid}  & \textbf{\% Succ} & \textbf{\% Opt} \\
\midrule
Zero-Shot             & 16& 0 & 0  \\
RAG-ICL [10k chars]   & 20& 6 & 6   \\
RAG-ICL [20k chars]   & 21& 9 & 9   \\
\midrule
ReAct    & 48& 41 & 37  \\
Self-Consistency ($k{=}5$)     & 59& 45 & 43  \\
Self-Refine ($k{=}5$)          & 51& 44 & 38  \\
\midrule
PTP\footnotemark/ L-ICL [$m=0$]    & 40& 33 & 28  \\
L-ICL [ours, $m=60$, 2k chars]   & \textbf{89}& \textbf{89} & \textbf{77} \\
\bottomrule
\end{tabular}
}
\caption{Performance on an 8$\times$8 two-room gridworld using DeepSeek V3. Paths start in one room and end in the other. \emph{Valid} plans never leave the grid or cross walls; \emph{Successful} plans reach their goals; and \emph{Optimal} plans are successful and use the minimum number of steps. L-ICL[$m$] denotes our method trained on $m$ examples, with the corresponding character count of L-ICL examples provided. All experiments are provided with an ASCII representation of the grid.}
\label{tab:teaser_gridworld8x8}
\end{table}

\footnotetext{PTP, introduced in~\citep{cohen2024watch} describes a method to prompt LLMs with partially specified programs}
\begin{figure*}[t]
  \centering
  \definecolor{liclnavy}{HTML}{243B53}
\definecolor{liclborder}{HTML}{7890A8}
\definecolor{liclpanel}{HTML}{E9F0F7}
\definecolor{licltext}{HTML}{1F2933}

\begin{tikzpicture}[
    font=\fontsize{9}{9.7}\selectfont,
    card/.style={
      rectangle,
      draw=liclborder,
      fill=liclpanel,
      rounded corners=1.5pt,
      line width=0.65pt,
      inner sep=0pt,
      outer sep=0pt,
      anchor=north west
    },
    stepnumber/.style={
      circle,
      fill=liclnavy,
      text=white,
      font=\fontsize{9}{9.5}\selectfont\bfseries,
      minimum size=4.8mm,
      inner sep=0pt
    },
    cardtitle/.style={
      anchor=west,
      text=liclnavy,
      font=\fontsize{9}{10}\selectfont\bfseries,
      inner sep=0pt
    },
    cardbody/.style={
      anchor=north west,
      align=left,
      text=licltext,
      font=\fontsize{9}{9.4}\selectfont,
      inner sep=0pt
    },
    codebody/.style={
      cardbody,
      font=\fontsize{9}{9.6}\selectfont\ttfamily,
      inner sep=0pt
    },
    forward/.style={
      -{Latex[length=1.6mm,width=1.15mm]},
      draw=liclnavy,
      line width=0.7pt
    },
    feedback/.style={
      -{Latex[length=1.7mm,width=1.2mm]},
      draw=liclnavy,
      line width=0.7pt,
      rounded corners=1.5mm
    },
    feedbacklabel/.style={
      text=liclnavy,
      font=\fontsize{9}{9.8}\selectfont\itshape,
      inner sep=0pt
    }
  ]
  \node[card, minimum width=4.70cm, minimum height=3.30cm] (prompt) at (0,0) {};
  \node[card, minimum width=4.95cm, minimum height=3.30cm] (trace) at (4.98,0) {};
  \node[card, minimum width=3.35cm, minimum height=3.30cm] (localize) at (10.21,0) {};
  \node[card, minimum width=3.80cm, minimum height=3.30cm] (doctest) at (13.84,0) {};

  \draw[draw=liclnavy, line width=1.1pt]
    ([xshift=1.5mm]prompt.north west) -- ([xshift=-1.5mm]prompt.north east);
  \draw[draw=liclnavy, line width=1.1pt]
    ([xshift=1.5mm]trace.north west) -- ([xshift=-1.5mm]trace.north east);
  \draw[draw=liclnavy, line width=1.1pt]
    ([xshift=1.5mm]localize.north west) -- ([xshift=-1.5mm]localize.north east);
  \draw[draw=liclnavy, line width=1.1pt]
    ([xshift=1.5mm]doctest.north west) -- ([xshift=-1.5mm]doctest.north east);

  \foreach \panel in {prompt,trace,localize,doctest}{
    \draw[draw=liclborder, line width=0.45pt]
      ([xshift=1.5mm,yshift=-7.1mm]\panel.north west) --
      ([xshift=-1.5mm,yshift=-7.1mm]\panel.north east);
  }

  \node[stepnumber] at ([xshift=4mm,yshift=-4.0mm]prompt.north west) {1};
  \node[cardtitle] at ([xshift=7.5mm,yshift=-4.0mm]prompt.north west) {Prompt template};
  \node[cardbody, font=\fontsize{9}{9}\selectfont, text width=4.26cm]
    at ([xshift=2.2mm,yshift=-7.6mm]prompt.north west) {%
      {\sffamily [ASCII GRID]}\\
      {\sffamily [PROGRAM FRAGMENT + TRACES]}\\
      \textbf{Subroutine docs (no code)}\\
      {\ttfamily get\_applicable\_actions}\\
      {\ttfamily get\_optimal\_actions}\\
      {\ttfamily at\_goal}\\
      \textbf{Q:} \texttt{((1,2),(6,5))}?
    };

  \node[stepnumber] at ([xshift=4mm,yshift=-4.0mm]trace.north west) {2};
  \node[cardtitle] at ([xshift=7.5mm,yshift=-4.0mm]trace.north west) {LLM-generated trace};
  \node[codebody, text width=4.51cm]
    at ([xshift=2.2mm,yshift=-8.2mm]trace.north west) {%
      >>> get\_applicable\_\\
      \phantom{>>> }actions((1,2))\\
      {}['move\_north',\\
      \phantom{[}\,'move\_east']\\
      ... many more calls ...\\
      >>> at\_goal((5,4),\\
      \phantom{>>> }(6,5)) -> False
    };

  \node[stepnumber] at ([xshift=4mm,yshift=-4.0mm]localize.north west) {3};
  \node[cardtitle] at ([xshift=7.5mm,yshift=-4.0mm]localize.north west) {Error Localization};
  \node[cardbody, text width=2.91cm]
    at ([xshift=2.2mm,yshift=-8.2mm]localize.north west) {%
      Verifier checks trace.\\[1.1mm]
      {\fontsize{9}{9.6}\selectfont\ttfamily get\_applicable\_}\\
      {\fontsize{9}{9.6}\selectfont\ttfamily actions((5,4))}
    };

  \node[stepnumber] at ([xshift=4mm,yshift=-4.0mm]doctest.north west) {4};
  \node[cardtitle] at ([xshift=7.5mm,yshift=-4.0mm]doctest.north west) {One-step doctest};
  \node[codebody, text width=3.36cm]
    at ([xshift=2.2mm,yshift=-8.2mm]doctest.north west) {%
      \rmfamily\fontsize{9}{9.4}\selectfont Correct output:\\[0.8mm]
      >>> get\_applicable\_\\
      \phantom{>>> }actions((5,4))\\
      {}['move\_east',\\
      \phantom{[}\,'move\_west']
    };

  \draw[forward] (prompt.east) -- (trace.west);
  \draw[forward] (trace.east) -- (localize.west);
  \draw[forward] (localize.east) -- (doctest.west);

  \draw[feedback]
    (doctest.south) -- ++(0,-4.8mm) -| (prompt.south);
  \node[feedbacklabel] at (8.82,-3.53)
    {insert localized doctest into the matching subroutine documentation};
\end{tikzpicture}
  \caption{The verifier finds the first failing trace call; L-ICL adds its corrected input--output pair as a one-step doctest.}
  \label{fig:arch-fig}
\end{figure*}
Table~\ref{tab:teaser_gridworld8x8} demonstrates this on a very simple 8$\times$8 two-room gridworld navigation task. Despite receiving complete information about the domain (grid layout and obstacles) no baseline method produces valid plans even 60\% of the time. Agentic and test-time-scaling approaches perform better, but still produce many invalid plans.
We conjecture that LLMs cannot build valid plans for this task because they \emph{fail to access the necessary domain-specific knowledge} in the prompt consistently.  This hypothesis is consistent with the failure of LLMs in these domains, and with their success in math and coding, where the necessary knowledge is {general}, and hence learnable in pre-training or fine-tuning.

In-context learning (ICL) is a natural remedy. However, complete solution trajectories demonstrate \emph{that} plans work, not \emph{why} individual steps are valid—leaving constraints implicit. As Table~\ref{tab:teaser_gridworld8x8} shows, even 20,000 characters of retrieved trajectories\footnote{RAG-ICL, retrieving demonstrations for tasks with similar start and end goals.} yield only 9\% success. The rules must still be inferred, and inference fails.


L-ICL escapes this by letting failures reveal which constraints need explicit specification. Rather than full trajectories, we augment prompts with \emph{localized} examples that demonstrate correct behavior on \emph{individual steps} where models err. We call this \textbf{Localized In-Context Learning} (L-ICL). This approach achieves higher performance with much less context: 2,000 characters of targeted corrections can outperform 20,000 characters of trajectories.  Generating L-ICL examples requires \emph{analyzing and correcting reasoning traces at training time}; we enable this by prompting models to produce structured reasoning traces, which are then corrected with a symbolic planner. While our primary experiments use symbolic correction outputs, we additionally evaluate controlled output corruption and corrections generated by a stronger LLM.. Thus, L-ICL might be viewed as distilling domain knowledge from a symbolic system into an LLM.  

Figure~\ref{fig:arch-fig} summarizes our approach, which builds on Program Trace Prompting (PTP)~\cite{cohen2024watch}. PTP recasts reasoning as producing a ``program trace'' for a partially specified program. A PTP prompt includes, for each type of reasoning ``step'', documentation (but not code) for a corresponding subroutine, along with (optional) example inputs and outputs.  For instance, a gridworld navigation task might include a subroutine \texttt{get\_applicable\_actions(cell)} that returns the set of obstacle-free cells adjacent to the input \texttt{cell}. Because \emph{no executable code is provided in PTP}, just documentation, the LLM must infer how to perform the reasoning step: e.g., in gridworld navigation, the LLM must infer which moves are valid for a task.  PTP's prompting scheme provides a natural insertion point for localized corrections: when a subroutine call fails, we locally augment that subroutine's documentation by adding a new input/output example. PTP thus acts as the limiting case of L-ICL when localized feedback is unavailable.\footnote{The input/output examples use Python's doctest syntax, a format well-represented in LLM training data, so readily understandable by code-trained LLMs.}

Given a planning task, we first prompt the LLM to generate a trace using the PTP format. We then analyze this trace programmatically to identify the \emph{first failing step}, i.e., the first subroutine call whose output violates domain constraints. An oracle (a symbolic simulator or verifier) provides the correct output for that input, yielding a localized correction. This correction is then inserted into the prompt.  For instance, if the LLM's first invalid move is from cell $(3,4)$, we L-ICL will add  to the prompt an example showing \texttt{get\_applicable\_actions((3,4))} should return \texttt{['move\_north', 'move\_south']}. This localized correction directly addresses the failure, and of course can also be generalized by the LLM to other similar cases.

This process iterates over multiple training instances, accumulating a bank of targeted examples that progressively refine the model's understanding of domain constraints. Crucially, the oracle is required only during training.

Experimentally, prompt augmentation with L-ICL dramatically reduces domain violations, and thus improves LLM planning performance across multiple domains. Beyond the results of Table~\ref{tab:teaser_gridworld8x8} and other gridworld tasks, we evaluate on classical planning benchmarks like BlocksWorld and Sokoban, seeing similar gains. 
L-ICL is also remarkably sample-efficient: peak performance is typically achieved with only 30--60 training examples.
L-ICL works across multiple LLMs (DeepSeek V3, DeepSeek V3.1, Claude Haiku 4.5, Claude Sonnet 4.5)~\citep{deepseek2024,deepseek2025v31,anthropic2025claude45,anthropic2025claudesonnet45}, and learned constraints can transfer across problem sizes (see Appendix~\ref{app:ood}).

To summarize our contributions:
(1) Using the PTP variant of semi-structured reasoning, we precisely measure constraint violation rates in LLM-generated plans across multiple planning domains, revealing that such violations are the dominant failure mode.
(2) We introduce L-ICL, a method that improves planning validity through localized, failure-driven corrections, and show that targeted examples outperform retrieval of complete trajectories even when the latter uses 10$\times$ more retrieved context.
(3) We demonstrate consistent improvements across multiple planning domains and four LLMs.

\section{Related Work}
\label{sec:related}

\subsection{LLM Planning: Capabilities and Limitations}

The planning capabilities of LLMs remain contested. One line of work reports strong performance on some planning tasks when LLMs are augmented with appropriate scaffolding. Tree of Thoughts explores branching
  reasoning trajectories \citep{yao2023tree}, RAP treats language-model reasoning as planning with a learned world model
  \citep{hao2023rap} and ReAct interleaves reasoning with actions
  \citep{yao2023react}
However, systematic evaluation on classical planning benchmarks reveals persistent failures. \citet{valmeekam2023planning} find that even capable LLMs struggle with International Planning Competition (IPC) domains
and \citet{stechly2024chain} further demonstrate that chain-of-thought improvements are brittle and fail to generalize beyond surface patterns. More broadly, reasoning performance can degrade sharply as problem complexity and trajectory length increase \citep{shojaee2025illusion}. These findings motivate the LLM-Modulo view, in which LLMs are treated as approximate knowledge sources whose proposals are checked by external verifiers, rather than as sound autonomous planners \citep{kambhampati2024position}.

We follow \citet{stechly2024chain} in using
structured reasoning traces to diagnose why LLM-generated
plans fail. However, instead of relying on a model fine-tuned to produce these traces, we use Program Trace Prompting (PTP) \citep{cohen2024watch}, which prompts LLMs with partially specified programs and asks them to produce Python-like execution traces. This makes intermediate computations parseable and provides natural insertion points for localized corrections, which free-form natural-language traces do not provide as consistently. We evaluate on symbolic planning benchmarks commonly used in the planning literature, including BlocksWorld, Gridworld, and Sokoban \citep{valmeekam2023planning,stechly2024chain,stechly2025on}, because these domains permit rigorous step-level verification.

\subsection{Approaches to Improve LLM Reasoning}

 \textbf{Structured Reasoning.} Chain-of-thought prompting elicits intermediate reasoning steps and can improve performance on many tasks \citep{wei2022cot}, although these explanations need not faithfully reflect the model's underlying computation \citep{turpin2023language}. Other approaches generate executable or partially executable programs: Program of Thoughts \citep{chen2022pot}, PAL \citep{gao2023pal}, and Chain of Code \citep{li2023chainofcode} delegate parts of the reasoning process to an external interpreter. PTP instead asks the model to predict the trace of a partially specified program. No executable implementations are provided, but the resulting structure makes individual reasoning steps observable and independently verifiable.
\par\noindent
\textbf{Test-Time Compute.} Several methods improve reasoning by expending more computation at inference. Self-Consistency~\citep{wang2023selfconsistency} aggregates multiple sampled paths; Tree of Thoughts~\citep{yao2023tree} explores branching reasoning trajectories; and Self-Refine~\citep{madaan2023selfrefine} iteratively improves outputs through self-critique.\footnote{Critically, \citet{stechly2025on} show that LLM self-verification is unreliable, making self-critique ineffective for planning.} ReAct~\citep{yao2023react} interleaves reasoning with environmental observation. In planning, CAPE uses action- precondition errors to re-prompt an LLM with corrective actions \citep{raman2024cape}. These methods repair or search for a solution to the current test problem and generally require multiple model calls or task-specific feedback at inference. These methods require multiple LLM calls or task-specific feedback at inference. In contrast, L-ICL accumulates corrections from training problems and requires only a single model call, with no verifier feedback, at test time. \par\noindent
\textbf{In-Context Learning.} ICL enables task adaptation through examples~\citep{brown2020language}, and is influenced by example selection~\citep{liu-etal-2022-makes} and format~\citep{min2022rethinking}. For planning, a natural approach is retrieving complete solution trajectories (RAG-ICL). However, we find this ineffective: 20,000 characters of retrieved trajectories yield only 9\% success on our gridworld benchmark. Complete trajectories demonstrate that solutions work but leave implicit why individual steps are valid. L-ICL addresses this by directly encoding constraints through localized input-output pairs.
\paragraph{Learning from mistakes.}
The intuition that step-level feedback can provide more informative credit assignment than final outcomes is well established. \citet{uesato2022solving} compare process- and outcome-based supervision for mathematical reasoning, while \citet{tyen2024llms} show that LLMs often fail to locate reasoning errors even when they can correct an error whose location is supplied. LEMA uses a stronger model to identify and correct erroneous reasoning steps and then fine-tunes models on the resulting correction data \citep{an2023learning}. L-ICL similarly uses localized corrections but retains them in the prompt rather than updating model weights.

Several methods also retain lessons from failures without fine-tuning. LEAP induces natural-language principles from mistakes on few-shot examples \citep{zhang2024context}, ExpeL extracts and recalls natural-language insights from success and failure experiences across training tasks \citep{zhao2024expel}, and Reflexion stores verbal reflections from previous interactions \citep{shinn2023reflexion}. These methods typically compress experience into general principles, insights, or reflections. L-ICL instead retains atomic, verifier-grounded input-output examples associated with specific reasoning subroutines. Thus, our contribution is not the general observation that localized feedback is useful, but a failure-driven method for constructing compact, reusable prompt-level supervision for planning.


\section{Method}
\label{sec:method}
\subsection{Background: Program Trace Prompting}
\label{sec:ptp}

Program Trace Prompting (PTP)~\citep{cohen2024watch} recasts reasoning as producing an execution trace for a partially specified program. A PTP prompt specifies each subroutine's signature and natural language documentation, provides a small number of example traces and the query problem. Crucially, the subroutine implementation is withheld - the LLM must therefore predict the output of the documented subroutines while producing a structured, parseable trace.\par\noindent
Across domains, we use a common planning interface containing subroutines that extract the initial and goal states, test goal completion, enumerate applicable actions, identify recommended/optimal actions, and apply an action. The interface is shared, but the state representation, action semantics and oracles are domain-specific. Because each intermediate call is explicit, PTP provides a natural location for inserting a correction when a call's predicted output is incorrect, ergo localized in-context learning.  Full subroutine specifications for each domain appear in Appendix~\ref{app:subroutines}.

\subsection{Localized In-Context Learning (L-ICL)}
\label{sec:licl}

The key insight behind L-ICL is that domain constraints can be taught more effectively through targeted examples than through complete solution trajectories. When an LLM violates a constraint (e.g., proposing to move through a wall), traditional approaches either reject the entire plan or provide feedback on the final outcome. L-ICL instead identifies the precise point of failure and injects a minimal correction for that specific subroutine call.\par\noindent
\textbf{First Failure Identification.} 
Given an LLM-generated trace, we parse its selected action sequence $a_1,\ldots,a_n$ and simulate that sequence from the initial state. Let $s_i$ be the state immediately before action $a_i$, and let $\mathcal{A}_{\mathrm{app}}(s_i)$ be the oracle-provided set of applicable actions. The first applicability error is $i^*_{\mathrm{app}}=\min\{i:a_i\notin\mathcal{A}_{\mathrm{app}}(s_i)\}$. Focusing on the first such error is deliberate. An invalid action renders the resulting state undefined, making later errors consequences of the invalid transition rather than independent failures. Correcting the root cause can thus prevent downstream errors.\par\noindent

In every domain, L-ICL localizes the first applicability  error. In Gridworld, Maze, Sokoban Grid, and Full Sokoban, it may also localize the first applicable but suboptimal action on the valid prefix preceding that invalid transition. A training problem can therefore contribute at most one correction of each enabled type, and actions following the first invalid transition are not interpreted. BlocksWorld uses applicability corrections only. We denote these fixed domain-level choices by $\Gamma_{\mathcal D}$ and provide the complete configuration and localization procedure in Appendix~\ref{app:correction_configuration}. \par \noindent
\textbf{Localized Correction.} For a localized error associated with subroutine $f$ and input $x$, we query the oracle to obtain the correct output $y^*=\mathcal{O}(f,x)$. This yields a correction tuple $(f,x,y^*)$. We format the correction as a doctest-style input-output example and insert it into the documentation for subroutine $f$, augmenting its original description with an additional example. This format, drawn from Python's widely used doctest convention, is well-represented in LLM training data. Appendix~\ref{app:correction_format} provides concrete examples. \par\noindent
\textbf{Iterative Accumulation.} 
L-ICL iterates over training problems $\{P_1,\ldots, P_m\}$. Problems are processed in batches using a prompt fixed at the beginning of each batch. After processing a batch, the newly collected corrections are merged with the existing correction library, and the prompt is rebuilt for the next batch. Prompt-visible corrections are keyed by their subroutine and exact input, so repeated failures on the same input produce only one rendered example. Algorithm~\ref{alg:licl} provides pseudocode. After training, the resulting prompt is fixed and
  evaluation requires neither oracle calls nor additional correction. L-ICL converges quickly: we see diminishing returns after only  30--60 training examples on our benchmark tasks (see Section~\ref{sec:results}).

\begin{algorithm}[t]
  \caption{Localized In-Context Learning (L-ICL)}
  \label{alg:licl}
  \small
  \begin{algorithmic}
  \REQUIRE Base prompt $\mathcal{P}_0$, training problems
  $\{P_1,\ldots,P_m\}$, oracle $\mathcal{O}$, domain correction
  configuration $\Gamma_{\mathcal D}$, and batch size $b$
  \ENSURE Augmented prompt $\mathcal{P}$
  \STATE $\mathcal{P}\gets\mathcal{P}_0$
  \STATE $\mathcal{C}\gets\emptyset$
  \hfill $\triangleright$ correction library
  \FOR{each batch $B$ of $b$ training problems}
      \STATE $\Delta\gets\emptyset$
      \FOR{$P_j\in B$}
          \STATE $\tau\gets
          \textsc{GenerateTrace}(\mathcal{P},P_j)$
          \STATE $E_j\gets
          \textsc{Localize}(\tau,\mathcal{O},\Gamma_{\mathcal D})$
          \FOR{each correction site $(f,x)\in E_j$}
              \STATE $y^*\gets\mathcal{O}(f,x)$
              \STATE $\Delta\gets
              \Delta\cup\{(f,x,y^*)\}$
          \ENDFOR
      \ENDFOR
      \STATE $\mathcal{C}\gets
      \textsc{MergeByInput}(\mathcal{C},\Delta)$
      \STATE $\mathcal{P}\gets
      \textsc{InsertCorrections}(\mathcal{P}_0,\mathcal{C})$
  \ENDFOR
  \STATE \textbf{return} $\mathcal{P}$
  \end{algorithmic}
  \end{algorithm}

\subsection{Experimental Domains}
\label{sec:domains}
We evaluate L-ICL on five symbolic planning domains: an 8$\times$8 two-room gridworld, a 10$\times$10 maze, a Sokoban-style gridworld with no pushable boxes, Full Sokoban with one movable box, and five-block BlocksWorld. Together, these domains span geometric and relational constraints, reversible and irreversible actions, and different state-tracking demands. Further details appear in Appendix~\ref{app:domains}.


\subsection{Baselines and Metrics}
\label{sec:baselines}
\label{sec:metrics}

We compare L-ICL against several approaches spanning prompting strategies, agentic methods, and retrieval. Detailed implementation appear in Appendix~\ref{app:baselines}, and prompts in Appendix~\ref{app:prompts}\par\noindent
\textbf{Zero-Shot.} The LLM receives the problem description and instructions only, measuring raw baseline capability.\par\noindent
\textbf{RAG-ICL.} We retrieve complete CoT-formatter solution trajectories for similar problems based on start/goal similarity, and evaluate at 10k and 20k character budgets.\par\noindent
\textbf{ReAct.} The LLM is instructed to interleave reasoning and action selection in its output. We evaluate a prompt-only version and an oracle-augmented version that queries a verifier during planning.\par\noindent
\textbf{Self-Consistency.} Majority voting with $k{=}5$ reasoning paths sampled at temperature 1.0.\par\noindent
\textbf{Self-Refine.} The LLM generates a  solution, then critiques and refines it, based on its own feedback, for $k{=}5$ iterations.\par\noindent
\textbf{Tree-of-Thoughts.} The LLM explores a tree of intermediate steps, evaluating and pruning branches (prompt-only, no external search).\par\noindent
Crucially, ReAct (Oracle) queries the verifier at \emph{test time} for each proposed action, while L-ICL uses the oracle \emph{only during training}. At inference, L-ICL requires a single forward pass with no external dependencies. For L-ICL, we report results with different numbers of training examples $m$ (denoted L-ICL[$m$]) to assess sample efficiency.

We evaluate plans along three axes that form a natural hierarchy. A plan is \emph{valid} if it violates no domain constraints (e.g., no wall collisions). A plan is \emph{successful} if it is valid and reaches the goal state. A plan is \emph{optimal} if it is successful and uses the minimum number of steps. Hence, a large valid-to-success gap indicates the model follows rules but fails to reach goals, and a large success-to-optimal gap indicates inefficient but functional plans.

\subsection{Experimental Setup}
\label{sec:setup}
Unless otherwise stated, the main domain comparison and learning curves use DeepSeek V3~\citep{deepseek2024}. The correction-selection, natural-language-correction, and feedback-robustness ablations use Claude Haiku 4.5~\citep{anthropic2025claude45} as the planning model, $m{=}60$ training problems, and 100 held-out 10$\times$10 Maze problems per condition. In the LLM-feedback condition, Claude Sonnet 4.6~\citep{anthropic2026claudesonnet46} supplies the correction outputs. A symbolic simulator continues to localize the first invalid action in all feedback-robustness conditions; only the source or reliability of the rendered correction output changes. In the model-specific learning curves, each model accumulates corrections from its own training failures. The frozen-prompt transfer experiment instead accumulates corrections using DeepSeek V3 and evaluates the resulting prompt on other models without further training.

For each domain, we generate 100 test problems with random start and goal configurations. For domains other than BlocksWorld, prompts use a textual state representation, as suggested in Figure~\ref{fig:arch-fig}, and, unless stated otherwise, an ASCII representation of the grid. Oracles are domain-specific: simple simulators for gridworlds and mazes, and Fast Downward \citep{helmert2006fast} and K-Star \citep{katz-lee-ijcai2023,Lee2023OnKS} for Sokoban and BlocksWorld. We use $T=1$ unless stated otherwise. Further details on datasets used are documented in Appendix~\ref{app:data}; the accompanying source archive is described in Appendix~\ref{app:code}.

\section{Results}
\label{sec:results}
We evaluate five questions: (1) Does L-ICL improve adherence to domain constraints? (2) Do failure-targeted corrections outperform untargeted or differently localized examples? (3) Does the correction mechanism depend on the PTP/doctest representation? (4) How efficiently and robustly does L-ICL use training feedback? (5) Do the gains persist across model families? $V$ denotes the percentage of generated plans containing no inapplicable action, while $S$ denotes the percentage of plans that are both valid and reach the goal. Additional ablations on correction type and frozen-prompt transfer appear in Appendix~\ref{app:mechanism_ablations}.

\subsection{L-ICL Learns Constraint Adherence}
\label{sec:results-main}

Table~\ref{tab:hero} presents our main results across all domains, with statistical significance tests for the main and ablation results appearing in Appendix~\ref{app:statistical-significance}. L-ICL consistently outperforms all baselines, often by substantial margins, addressing several aspects of planning.

\begin{table*}[t]
\centering
{
\small
\setlength{\tabcolsep}{1mm}
\begin{tabular}{@{}l cc cc cc cc cc@{}}
\toprule
& \multicolumn{2}{c}{\textbf{8$\times$8 Grid}} & \multicolumn{2}{c}{\textbf{10$\times$10 Maze}} & \multicolumn{2}{c}{\textbf{Sokoban Grid}} & \multicolumn{2}{c}{\textbf{Full Sokoban}} & \multicolumn{2}{c}{\textbf{BlocksWorld}} \\
\textbf{Method} & V & S & V & S & V & S & V & S & V & S \\
\midrule
Zero-Shot & 16& 0 & 3& 0 & 15& 0 & 1& 0 & 10& 10 \\
RAG-ICL (10k chars) & 20& 6 & 7& 1 & 17& 4 & 31& 11& 25& 25 \\
RAG-ICL (20k chars) & 21& 9 & 7& 4 & 25& 10 & 36& 15& 32& 32 \\
\midrule
ReAct (Prompt-Only) & 48& 41 & 6& 5 & 19& 12 & 1& 0 & 46& 45 \\
Self-Consistency ($k{=}5$) & 59& 45 & 3& 3 & 10& 5 & 2& 1 & 31& 31\\
Self-Refine ($k{=}5$) & 51& 44 & 3& 1 & 13& 8 & 0& 0 & 49& 49 \\
ToT (Prompt-Only) & 33& 12 & 1& 0 & 3& 2 & 0& 0 & 50& 40 \\
\midrule
\textit{ReAct (Oracle f/b)}$^\dagger$ & 55& 45 & 6& 5 & 21& 13 & 3& 0 & \underline{51}& \underline{51} \\
\midrule
L-ICL[$m{=}0$] (ours) & 40& 33 & 20& 16 & 21& 17 & 19& 13 & 50& 48 \\
L-ICL[$m{=}60$] (ours) & \textbf{89}& \textbf{89} & \underline{40}& \underline{21} & \textbf{63}& \textbf{49}& \textbf{46}& \textbf{20} & \textbf{68}& \textbf{66}\\
\midrule
  L-ICL[$m{=}0$]$^*$ (ours) & 19& 12 & 7& 6 & 10& 8 & 12& 9 &
  -- & -- \\
  L-ICL[$m{=}60$]$^*$ (ours) & \underline{73}& \underline{63} &
  \textbf{57}& \textbf{27} & \underline{62}& \underline{44}&
  \underline{42}& \underline{14} & --& -- \\
\bottomrule
\end{tabular}
}
\caption{Main results using DeepSeek V3 and 100 held-out problems per condition. We report \%(V)alid and \%(S)uccessful. All baselines receive ASCII grid representations. L-ICL[$m$] denotes training on $m$ examples. Best results in bold, second-best underlined. Blocksworld has no grid representation, so the starred condition is inapplicable.$\dagger$ReAct (Oracle f/b) receives oracle feedback at inference time. $^*$L-ICL (no grid) methods receive no ASCII grid, relying on L-ICL to infer structure.}
\label{tab:hero}
\end{table*}

\textbf{Constraint learning is consistent across domains.} At $m{=}60$, primary-condition validity reaches 89\% on the 8$\times$8 Grid, 40\% on the 10$\times$10 Maze, 63\% on Sokoban Grid, 46\% on Full Sokoban, and 68\% on BlocksWorld. These gains occur for geometric constraints, irreversible box-pushing constraints, and relational BlocksWorld preconditions, indicating that L-ICL is not tied to a particular action representation. BlocksWorld is especially informative because it uses applicability feedback alone.

\textbf{Validity and goal-directed planning remain separable.} The gaps between validity and success at $m{=}60$ are 0 points on the 8$\times$8 Grid, 19 on the Maze, 14 on Sokoban Grid, 26 on Full Sokoban, and 2 on BlocksWorld. Thus, L-ICL most consistently improves  action applicability, which serves as the primary outcome we analyze, while longer-horizon coordination remains difficult in Maze and Full Sokoban. Appendix~\ref{app:valid_success_gap} analyzes this distinction further.

\begin{figure}[t]
  \centering
  \includegraphics[width=\columnwidth]{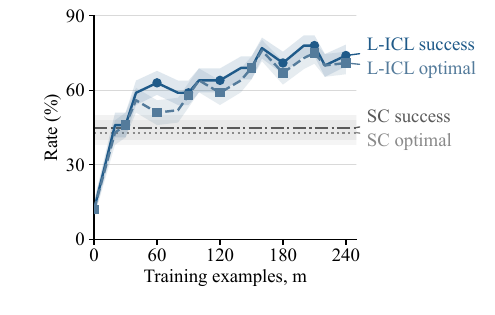}
  \caption{8$\times$8 Gridworld learning curves. Success and Optimal rates vs.\ training examples. L-ICL (without the ASCII grid) improves rapidly in the first 30--60 examples, substantially outperforming all baselines, which are given access to the ASCII grid (horizontal line shows best baseline). Shaded bands show $\pm$ 1 binomial standard error.}
  \label{fig:gridworld_curves}
\end{figure}



\subsection{L-ICL Tolerates Imperfect Correction Feedback}
\label{sec:results-imperfect-feedback}

We test whether L-ICL requires correction outputs from a perfect symbolic oracle in two complementary ways. First, we randomly corrupt a controlled fraction of the symbolic oracle's correction outputs. Second, we replace the oracle with predictions from a stronger LLM, Claude Sonnet 4.6, with full access to the ASCII grid. All conditions use Claude Haiku 4.5 as the planning model, $m{=}60$ training problems, an ASCII grid, and the same 100 held-out 10$\times$10 Maze problems.  Performance declines as correction reliability decreases but does not collapse. With 50\% randomly corrupted outputs, L-ICL retains $36/57\approx63\%$ of its clean validity and $35/55\approx64\%$ of its clean success. Sonnet-generated corrections yield 35\% validity and success, roughly comparable to the heavily corrupted symbolic condition. These results show that L-ICL can extract useful signal from imperfect correction outputs, while the gap from clean supervision confirms that correction quality remains important.

  \begin{table}[t]
  \centering
  
  \small
  \setlength{\tabcolsep}{2.5mm}
  \begin{tabular}{@{}lcc@{}}
  \toprule
  \textbf{Correction-output source} & \textbf{V} & \textbf{S} \\
  \midrule
  Symbolic oracle (clean)       & \textbf{57} & \textbf{55} \\
  Symbolic $+$ 10\% corruption & 51 & 48 \\
  Symbolic $+$ 20\% corruption & 47 & 46 \\
  Symbolic $+$ 50\% corruption & 36 & 35 \\
  \midrule
  Claude Sonnet 4.6             & 35 & 35 \\
  \bottomrule
  \end{tabular}
  
  \caption{Robustness to imperfect correction feedback on the
  10$\times$10 Maze. Claude Haiku 4.5 is the planning model in every
  row; Claude Sonnet 4.6 supplies the correction outputs in the final
  row. Each condition uses $m{=}60$ training problems and 100 held-out
  test problems.}
  \label{tab:imperfect_feedback}
  \end{table}

\subsection{L-ICL Is More Efficient Than Retrieval-Based ICL}
\label{sec:results-efficiency}

\begin{figure}[t]
  \centering
  \includegraphics[width=\columnwidth]{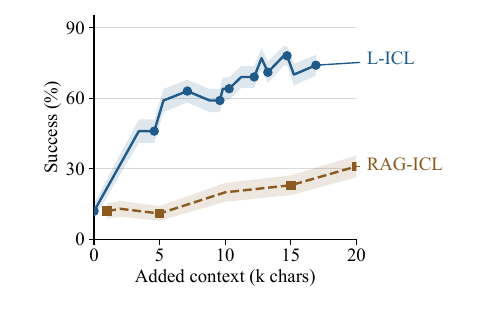}
  \caption{Sample efficiency: L-ICL vs.\ RAG-ICL. Success rate vs.\ context size (characters) on 8$\times$8 Gridworld. L-ICL achieves higher performance with substantially less context.}
  \label{fig:efficiency}
\end{figure}

Figure~\ref{fig:efficiency} compares the amount of additional demonstration context used by L-ICL and RAG-ICL. Approximately 2,000 characters of localized corrections already outperform 20,000 characters of retrieved trajectories, a tenfold reduction in added context. This supports a failure-driven selection principle: complete trajectories devote most of their tokens to steps the model may already handle, whereas L-ICL spends context on inputs where the model demonstrably failed.

\subsection{L-ICL Does Not Need an ASCII Grid}
\label{sec:results-grid-ablation}

\begin{figure}[t]
  \centering
  \includegraphics[width=\columnwidth]{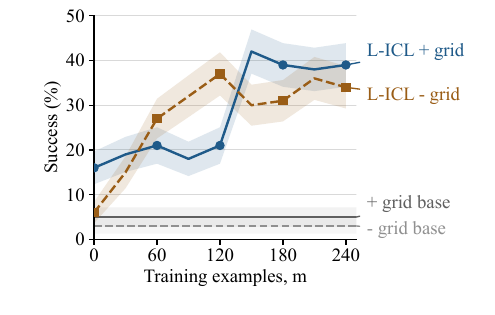}
  \caption{Grid ablation on 10$\times$10 Maze. The ASCII grid accelerates early learning but does not change peak performance. Without L-ICL, the grid provides little benefit.}
  \label{fig:grid_ablation}
\end{figure}

For spatial tasks, prompts normally include an ASCII grid. We therefore test whether L-ICL still helps without this visualization. Figure~\ref{fig:grid_ablation} shows the learning curve for L-ICL on the 10x10 grid task with and without the ASCII grid. 
We find that the grid improves early success (21\% at $m{=}30$ with grid vs.\ 15\% without), but not peak performance (39\% vs.\ 37\%). Thus, L-ICL does not require visual scaffolding for peak performance, although it can accelerate early learning. 

\subsection{L-ICL is LLM-agnostic}
\label{sec:results-llm-ablation}
\begin{figure}[t]
  \centering
  \includegraphics[width=\columnwidth]{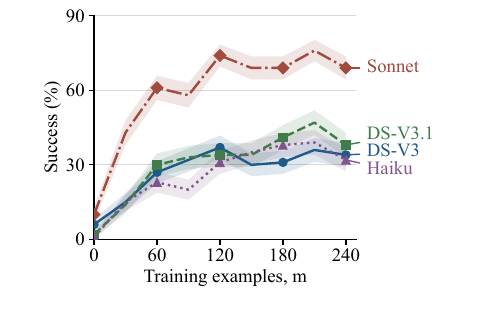}
  \caption{L-ICL across LLMs. Success rate on 10$\times$10 Maze for four models. All improve substantially; Claude Sonnet 4.5 shows the largest gains (10\% $\to$ 76\%).}
  \label{fig:llm_ablation}
\end{figure}

Figure~\ref{fig:llm_ablation} evaluates the same no-grid 10$\times$10 Maze procedure on DeepSeek V3 and V3.1~\citep{deepseek2024,deepseek2025v31}, Claude Haiku 4.5 and Sonnet 4.5~\citep{anthropic2025claude45,anthropic2025claudesonnet45}, with each model accumulating corrections from its own training failures. The magnitude and learning dynamics differ, but all four models benefit greatly. Appendix~\ref{app:frozen_prompt_transfer} additionally evaluates a single frozen DeepSeek-V3 correction prompt on the Claude models, showing strong transfer of L-ICL performance, regardless of generating model.

\section{Discussion}
\label{sec:discussion}

Our experiments show that L-ICL improves the validity of LLM-generated plans across all five evaluated domains; across the domain suite, improved action applicability is L-ICL's most consistent effect. Beyond the aggregate comparison, the results clarify both the primary effect of localized corrections and the aspects of planning that remain difficult.

\subsection{L-ICL as In-Context Unit Testing}
\label{sec:discussion-unit-tests}

In software engineering, unit tests specify the desired behavior of an individual component on particular inputs. L-ICL corrections are analogous in granularity: each correction specifies the desired output of one reasoning subroutine at an input where the model failed. The analogy is not exact---unit tests evaluate executable code, whereas L-ICL conditions a model---but it captures how L-ICL concentrates supervision on individual reasoning steps. Full-trajectory demonstrations instead describe many interacting steps at once. Such demonstrations can illustrate an overall solution process, but much of their context may be devoted to steps that the model already performs correctly. The efficiency results imply that localized corrections allocate limited context more effectively by concentrating examples at observed failure sites.

\subsection{Qualitative Evidence: Guessing to Navigation}
\label{sec:discussion-qualitative}

\begin{figure}[t]
    \centering
    \includegraphics[width=0.48\columnwidth]
    {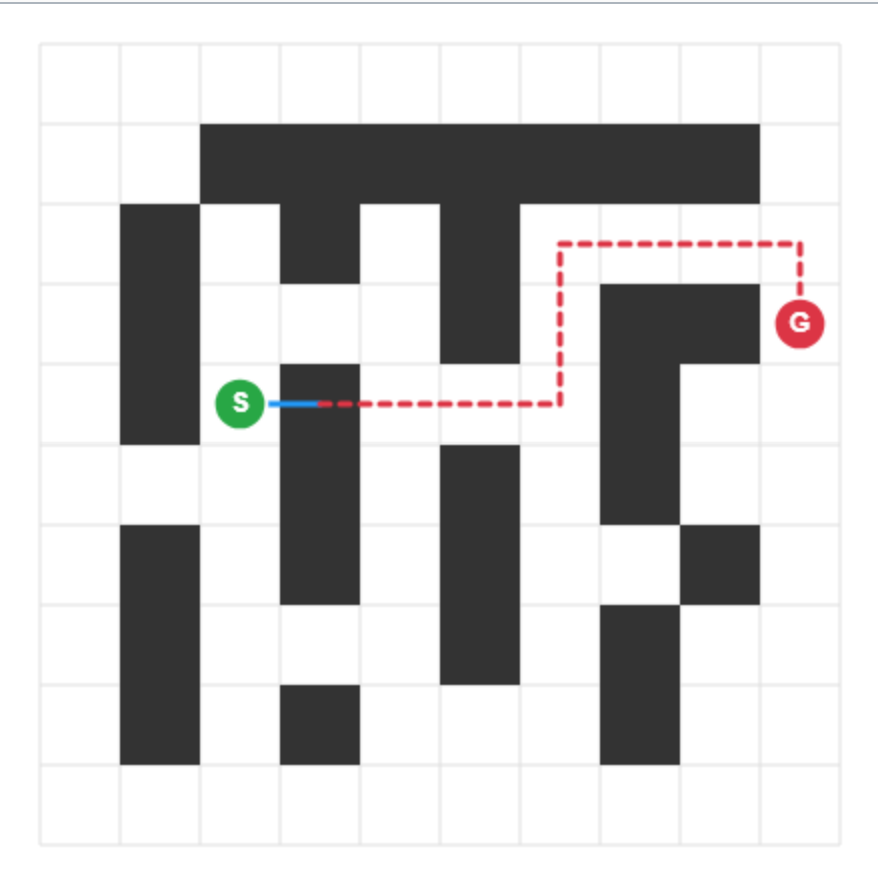}\hfill
    \includegraphics[width=0.48\columnwidth]
    {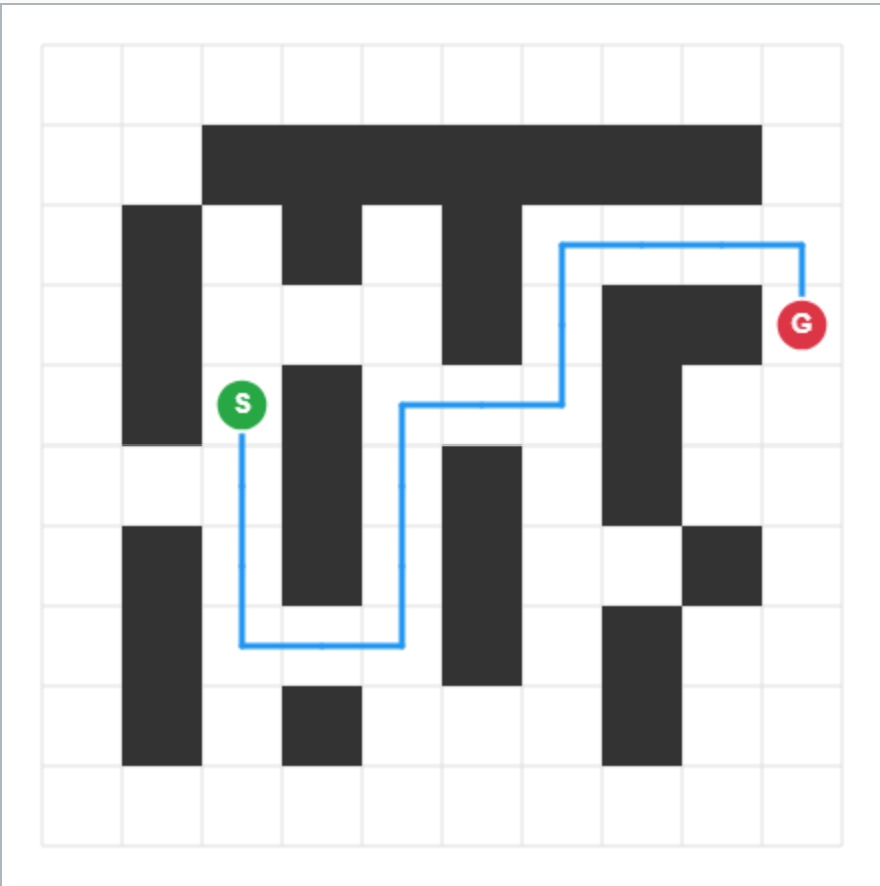}
    \caption{Qualitative no-grid example on the same held-out maze
    as the number of training problems increases. At $m{=}0$ (left), the
    trajectory crosses walls; at $m{=}60$ (right), it reaches
    the goal using only applicable moves.}
    \label{fig:m_qualitative_maze}
  \end{figure}

Figure~\ref{fig:m_qualitative_maze} illustrates the behavioral change associated with L-ICL. At $m{=}0$, the generated trajectory quickly attempts an inapplicable move through a wall. At $m{=}60$, the model produces a coherent start-to-goal trajectory that respects the maze constraints. As the task still provides its constraints textually, this example demonstrates the use of textual constraint information rather than an abstract transferable prior.

\subsection{Limitations and Scope}
\label{sec:limitations}

We focus on classical planning because it permits exact step-level diagnosis of constraint violations. Our five domains span geometric and relational constraints, but do not establish transfer to open-ended tasks with ambiguous states or action preconditions; we leave that extension to future work. L-ICL still requires training-time supervision, although inference uses a fixed prompt without oracle calls. Our robustness experiments show that L-ICL tolerates corrupted or LLM-generated correction outputs, but as this remains a controlled Maze experiment, the reliability of learned supervision in less structured domains requires further study. Finally, valid actions do not guarantee a goal-reaching plan: validity-to-success gaps suggest difficulty with long-horizon composition in Maze and multi-object coordination in Sokoban, but do not isolate the failure source. Combining L-ICL with search or value-guided action selection is a natural extension.
\section{Conclusion}
\label{sec:conclusion}

LLMs frequently violate domain constraints even when those are explicitly supplied in the prompt. L-ICL addresses this failure by localizing errors on training problems and accumulating compact input-output corrections at the corresponding reasoning subroutines. Across five domains, the resulting prompts improve action applicability without inference-time oracle feedback; on the 8$\times$8 Gridworld, L-ICL reaches 89\% validity versus 59\% for the strongest baseline.
The results also demonstrate the value of allocating prompt context according to observed failures. In the 8$\times$8 comparison, roughly 2,000 characters of localized corrections perform better than 20,000 characters of retrieved trajectories. This supports failure-targeted correction as a context-efficient alternative to supplying full solutions. L-ICL does not by itself solve planning: a sequence of applicable actions may still terminate before the goal or make choices from which recovery is difficult. Instead, L-ICL can serve as a procedural hardening component that makes local constraint violations less frequent and can be combined with methods for longer-horizon search and action selection. More broadly, the results suggest that structured, failure-localized examples can improve an LLM's use of provided knowledge in task specifications.
\clearpage
\subsection*{Acknowledgements}
The authors thank Chenyan Xiong for his guidance and suggestions,  Cassandra A. Cohen for helping to develop and support the PTP infrastructure, and Google Inc. for a research grant that partially supported this work.
{\small
\bibliography{example_paper}

@inproceedings{valmeekam2023planning,
  title={On the Planning Abilities of Large Language Models---A Critical Investigation},
  author={Valmeekam, Karthik and Marquez, Matthew and Sreedharan, Sarath and Kambhampati, Subbarao},
  booktitle={Advances in Neural Information Processing Systems},
  volume={36},
  pages={75993--76005},
  doi={10.52202/075280-3320},
  url={https://proceedings.neurips.cc/paper_files/paper/2023/hash/efb2072a358cefb75886a315a6fcf880-Abstract-Conference.html},
  year={2023}
}

@inproceedings{kambhampati2024position,
  title={Position: {LLM}s Can't Plan, But Can Help Planning in {LLM}-Modulo Frameworks},
  author={Kambhampati, Subbarao and Valmeekam, Karthik and Guan, Lin and Verma, Mudit and Stechly, Kaya and Bhambri, Siddhant and Saldyt, Lucas Paul and Murthy, Anil B.},
  booktitle={Proceedings of the 41st International Conference on Machine Learning},
  series={Proceedings of Machine Learning Research},
  volume={235},
  pages={22895--22907},
  publisher={PMLR},
  year={2024},
  url={https://proceedings.mlr.press/v235/kambhampati24a.html}
}

@inproceedings{stechly2024chain,
  title={Chain of Thoughtlessness? An Analysis of {CoT} in Planning},
  author={Stechly, Kaya and Valmeekam, Karthik and Kambhampati, Subbarao},
  booktitle={Advances in Neural Information Processing Systems},
  volume={37},
  pages={29106--29141},
  doi={10.52202/079017-0917},
  url={https://proceedings.neurips.cc/paper_files/paper/2024/hash/3365d974ce309623bd8151082d78206c-Abstract-Conference.html},
  year={2024}
}

@inproceedings{
stechly2025on,
title={On the self-verification limitations of large language models on reasoning and planning tasks},
author={Kaya Stechly and Karthik Valmeekam and Subbarao Kambhampati},
booktitle={The Thirteenth International Conference on Learning Representations},
year={2025},
url={https://openreview.net/forum?id=4O0v4s3IzY}
}

@inproceedings{shojaee2025illusion,
  title={The Illusion of Thinking: Understanding the Strengths and Limitations of Reasoning Models via the Lens of Problem Complexity},
  author={Shojaee, Parshin and Mirzadeh, Iman and {Alizadeh Vahid}, Keivan and Horton, Maxwell and Bengio, Samy and Farajtabar, Mehrdad},
  booktitle={Advances in Neural Information Processing Systems},
  volume={38},
  pages={108018--108059},
  url={https://proceedings.neurips.cc/paper_files/paper/2025/hash/9b26ad15462c81548c0689188d2e8018-Abstract-Conference.html},
  year={2025}
}

@inproceedings{raman2024cape,
  title={{CAPE}: Corrective Actions from Precondition Errors using Large Language Models},
  author={{Sundara Raman}, Shreyas and Cohen, Vanya and Idrees, Ifrah and Rosen, Eric and Mooney, Raymond and Tellex, Stefanie and Paulius, David},
  booktitle={2024 IEEE International Conference on Robotics and Automation (ICRA)},
  pages={14070--14077},
  year={2024},
  publisher={IEEE},
  doi={10.1109/ICRA57147.2024.10611376},
  url={https://doi.org/10.1109/ICRA57147.2024.10611376}
}

@inproceedings{yao2023tree,
  title={Tree of Thoughts: Deliberate Problem Solving with Large Language Models},
  author={Yao, Shunyu and Yu, Dian and Zhao, Jeffrey and Shafran, Izhak and Griffiths, Tom and Cao, Yuan and Narasimhan, Karthik},
  booktitle={Advances in Neural Information Processing Systems},
  volume={36},
  pages={11809--11822},
  doi={10.52202/075280-0517},
  url={https://proceedings.neurips.cc/paper_files/paper/2023/hash/271db9922b8d1f4dd7aaef84ed5ac703-Abstract-Conference.html},
  year={2023}
}

@inproceedings{yao2023react,
  title={ReAct: Synergizing Reasoning and Acting in Language Models},
  author={Yao, Shunyu and Zhao, Jeffrey and Yu, Dian and Du, Nan and Shafran, Izhak and Narasimhan, Karthik and Cao, Yuan},
  booktitle={The Eleventh International Conference on Learning Representations},
  year={2023},
  url={https://openreview.net/forum?id=WE_vluYUL-X}
}

@inproceedings{hao2023rap,
  title={Reasoning with Language Model is Planning with World Model},
  author={Hao, Shibo and Gu, Yi and Ma, Haodi and Hong, Joshua and Wang, Zhen and Wang, Daisy and Hu, Zhiting},
  booktitle={Proceedings of the 2023 Conference on Empirical Methods in Natural Language Processing},
  pages={8154--8173},
  year={2023},
  publisher={Association for Computational Linguistics},
  doi={10.18653/v1/2023.emnlp-main.507},
  url={https://aclanthology.org/2023.emnlp-main.507/}
}

@inproceedings{wei2022cot,
  title={Chain-of-Thought Prompting Elicits Reasoning in Large Language Models},
  author={Wei, Jason and Wang, Xuezhi and Schuurmans, Dale and Bosma, Maarten and Ichter, Brian and Xia, Fei and Chi, Ed H. and Le, Quoc V. and Zhou, Denny},
  booktitle={Advances in Neural Information Processing Systems},
  volume={35},
  pages={24824--24837},
  doi={10.52202/068431-1800},
  url={https://proceedings.neurips.cc/paper_files/paper/2022/hash/9d5609613524ecf4f15af0f7b31abca4-Abstract-Conference.html},
  year={2022}
}

@misc{cohen2024watch,
  title={Watch Your Steps: Observable and Modular Chains of Thought},
  author={Cohen, Cassandra A and Cohen, William W},
  year={2024},
  eprint={2409.15359},
  archivePrefix={arXiv},
  primaryClass={cs.CL},
  doi={10.48550/arXiv.2409.15359},
  url={https://arxiv.org/abs/2409.15359}
}

@article{chen2022pot,
  title={Program of Thoughts Prompting: Disentangling Computation from Reasoning for Numerical Reasoning Tasks},
  author={Chen, Wenhu and Ma, Xueguang and Wang, Xinyi and Cohen, William W},
  journal={Transactions on Machine Learning Research},
  year={2023},
  issn={2835-8856},
  url={https://openreview.net/forum?id=YfZ4ZPt8zd}
}

@inproceedings{gao2023pal,
  title={{PAL}: Program-aided Language Models},
  author={Gao, Luyu and Madaan, Aman and Zhou, Shuyan and Alon, Uri and Liu, Pengfei and Yang, Yiming and Callan, Jamie and Neubig, Graham},
  booktitle={Proceedings of the 40th International Conference on Machine Learning},
  series={Proceedings of Machine Learning Research},
  volume={202},
  pages={10764--10799},
  publisher={PMLR},
  year={2023},
  url={https://proceedings.mlr.press/v202/gao23f.html}
}

@inproceedings{li2023chainofcode,
  title={Chain of Code: Reasoning with a Language Model-Augmented Code Emulator},
  author={Li, Chengshu and Liang, Jacky and Zeng, Andy and Chen, Xinyun and Hausman, Karol and Sadigh, Dorsa and Levine, Sergey and Fei-Fei, Li and Xia, Fei and Ichter, Brian},
  booktitle={Proceedings of the 41st International Conference on Machine Learning},
  series={Proceedings of Machine Learning Research},
  volume={235},
  pages={28259--28277},
  publisher={PMLR},
  year={2024},
  url={https://proceedings.mlr.press/v235/li24ar.html}
}

@inproceedings{wang2023selfconsistency,
  title={Self-Consistency Improves Chain of Thought Reasoning in Language Models},
  author={Wang, Xuezhi and Wei, Jason and Schuurmans, Dale and Le, Quoc V. and Chi, Ed H. and Narang, Sharan and Chowdhery, Aakanksha and Zhou, Denny},
  booktitle={The Eleventh International Conference on Learning Representations},
  year={2023},
  url={https://openreview.net/forum?id=1PL1NIMMrw}
}

@inproceedings{madaan2023selfrefine,
  title={Self-Refine: Iterative Refinement with Self-Feedback},
  author={Madaan, Aman and Tandon, Niket and Gupta, Prakhar and Hallinan, Skyler and Gao, Luyu and Wiegreffe, Sarah and Alon, Uri and Dziri, Nouha and Prabhumoye, Shrimai and Yang, Yiming and Gupta, Shashank and Majumder, Bodhisattwa Prasad and Hermann, Katherine and Welleck, Sean and Yazdanbakhsh, Amir and Clark, Peter},
  booktitle={Advances in Neural Information Processing Systems},
  volume={36},
  pages={46534--46594},
  doi={10.52202/075280-2019},
  url={https://proceedings.neurips.cc/paper_files/paper/2023/hash/91edff07232fb1b55a505a9e9f6c0ff3-Abstract-Conference.html},
  year={2023}
}

@inproceedings{tyen2024llms,
  title={{LLM}s cannot find reasoning errors, but can correct them given the error location},
  author={Tyen, Gladys and Mansoor, Hassan and Carbune, Victor and Chen, Peter and Mak, Tony},
  booktitle={Findings of the Association for Computational Linguistics: ACL 2024},
  pages={13894--13908},
  year={2024},
  publisher={Association for Computational Linguistics},
  doi={10.18653/v1/2024.findings-acl.826},
  url={https://aclanthology.org/2024.findings-acl.826/}
}

@misc{uesato2022solving,
  title={Solving Math Word Problems with Process- and Outcome-Based Feedback},
  author={Uesato, Jonathan and Kushman, Nate and Kumar, Ramana and Song, Francis and Siegel, Noah and Wang, Lisa and Creswell, Antonia and Irving, Geoffrey and Higgins, Irina},
  year={2022},
  eprint={2211.14275},
  archivePrefix={arXiv},
  primaryClass={cs.LG},
  doi={10.48550/arXiv.2211.14275},
  url={https://arxiv.org/abs/2211.14275}
}

@inproceedings{brown2020language,
  title={Language Models are Few-Shot Learners},
  author={Brown, Tom and Mann, Benjamin and Ryder, Nick and Subbiah, Melanie and Kaplan, Jared D and Dhariwal, Prafulla and Neelakantan, Arvind and Shyam, Pranav and Sastry, Girish and Askell, Amanda and Agarwal, Sandhini and Herbert-Voss, Ariel and Krueger, Gretchen and Henighan, Tom and Child, Rewon and Ramesh, Aditya and Ziegler, Daniel and Wu, Jeffrey and Winter, Clemens and Hesse, Chris and Chen, Mark and Sigler, Eric and Litwin, Mateusz and Gray, Scott and Chess, Benjamin and Clark, Jack and Berner, Christopher and McCandlish, Sam and Radford, Alec and Sutskever, Ilya and Amodei, Dario},
  booktitle={Advances in Neural Information Processing Systems},
  volume={33},
  pages={1877--1901},
  publisher={Curran Associates, Inc.},
  url={https://proceedings.neurips.cc/paper/2020/hash/1457c0d6bfcb4967418bfb8ac142f64a-Abstract.html},
  year={2020}
}

@inproceedings{liu-etal-2022-makes,
    title = "What Makes Good In-Context Examples for {GPT}-3?",
    author = "Liu, Jiachang  and
      Shen, Dinghan  and
      Zhang, Yizhe  and
      Dolan, Bill  and
      Carin, Lawrence  and
      Chen, Weizhu",
    editor = "Agirre, Eneko  and
      Apidianaki, Marianna  and
      Vuli{\'c}, Ivan",
    booktitle = "Proceedings of Deep Learning Inside Out (DeeLIO 2022): The 3rd Workshop on Knowledge Extraction and Integration for Deep Learning Architectures",
    month = may,
    year = "2022",
    address = "Dublin, Ireland and Online",
    publisher = "Association for Computational Linguistics",
    url = "https://aclanthology.org/2022.deelio-1.10/",
    doi = "10.18653/v1/2022.deelio-1.10",
    pages = "100--114"
}

@inproceedings{min2022rethinking,
  title={Rethinking the Role of Demonstrations: What Makes In-Context Learning Work?},
  author={Min, Sewon and Lyu, Xinxi and Holtzman, Ari and Artetxe, Mikel and Lewis, Mike and Hajishirzi, Hannaneh and Zettlemoyer, Luke},
  booktitle={Proceedings of the 2022 Conference on Empirical Methods in Natural Language Processing},
  pages={11048--11064},
  year={2022},
  address={Abu Dhabi, United Arab Emirates},
  publisher={Association for Computational Linguistics},
  doi={10.18653/v1/2022.emnlp-main.759},
  url={https://aclanthology.org/2022.emnlp-main.759/}
}

@inproceedings{turpin2023language,
  title={Language Models Don't Always Say What They Think: Unfaithful Explanations in Chain-of-Thought Prompting},
  author={Turpin, Miles and Michael, Julian and Perez, Ethan and Bowman, Samuel R.},
  booktitle={Advances in Neural Information Processing Systems},
  volume={36},
  pages={74952--74965},
  doi={10.52202/075280-3275},
  url={https://proceedings.neurips.cc/paper_files/paper/2023/hash/ed3fea9033a80fea1376299fa7863f4a-Abstract-Conference.html},
  year={2023}
}

@inproceedings{zhang2024context,
  title={In-Context Principle Learning from Mistakes},
  author={Zhang, Tianjun and Madaan, Aman and Gao, Luyu and Zheng, Steven and Mishra, Swaroop and Yang, Yiming and Tandon, Niket and Alon, Uri},
  booktitle={Proceedings of the 41st International Conference on Machine Learning},
  series={Proceedings of Machine Learning Research},
  volume={235},
  pages={59520--59558},
  publisher={PMLR},
  year={2024},
  url={https://proceedings.mlr.press/v235/zhang24at.html}
}

@article{zhao2024expel,
  title={{ExpeL}: {LLM} Agents Are Experiential Learners},
  author={Zhao, Andrew and Huang, Daniel and Xu, Quentin and Lin, Matthieu and Liu, Yong-Jin and Huang, Gao},
  journal={Proceedings of the AAAI Conference on Artificial Intelligence},
  volume={38},
  number={17},
  pages={19632--19642},
  year={2024},
  doi={10.1609/aaai.v38i17.29936},
  url={https://ojs.aaai.org/index.php/AAAI/article/view/29936}
}

@misc{an2023learning,
  title={Learning from Mistakes Makes {LLM} Better Reasoner},
  author={An, Shengnan and Ma, Zexiong and Lin, Zeqi and Zheng, Nanning and Lou, Jian-Guang and Chen, Weizhu},
  year={2023},
  eprint={2310.20689},
  archivePrefix={arXiv},
  primaryClass={cs.CL},
  doi={10.48550/arXiv.2310.20689},
  url={https://arxiv.org/abs/2310.20689}
}

@inproceedings{lehnert2024beyond,
  title={Beyond {A*}: Better Planning with Transformers via Search Dynamics Bootstrapping},
  author={Lehnert, Lucas and Sukhbaatar, Sainbayar and Su, DiJia and Zheng, Qinqing and McVay, Paul and Rabbat, Michael and Tian, Yuandong},
  booktitle={First Conference on Language Modeling},
  year={2024},
  url={https://openreview.net/forum?id=SGoVIC0u0f}
}

@article{helmert2006fast,
  title={The Fast Downward Planning System},
  author={Helmert, Malte},
  journal={Journal of Artificial Intelligence Research},
  volume={26},
  pages={191--246},
  year={2006},
  doi={10.1613/jair.1705},
  url={https://doi.org/10.1613/jair.1705}
}

@article{Lee2023OnKS,
  title={On {K*} Search for Top-K Planning},
  author={Lee, Junkyu and Katz, Michael and Sohrabi, Shirin},
  journal={Proceedings of the International Symposium on Combinatorial Search},
  volume={16},
  number={1},
  pages={38--46},
  year={2023},
  publisher={AAAI Press},
  doi={10.1609/socs.v16i1.27281},
  url={https://ojs.aaai.org/index.php/SOCS/article/view/27281}
}

@inproceedings{katz-lee-ijcai2023,
  author={Katz, Michael and Lee, Junkyu},
  title={{K\ensuremath{*} Search over Orbit Space for Top-k Planning}},
  booktitle={Proceedings of the Thirty-Second International Joint Conference on Artificial Intelligence},
  pages={5368--5376},
  publisher={International Joint Conferences on Artificial Intelligence Organization},
  year={2023},
  doi={10.24963/ijcai.2023/596},
  url={https://www.ijcai.org/proceedings/2023/596}
}

@misc{deepseek2024,
  title={{DeepSeek-V3} Technical Report},
  author={{DeepSeek-AI}},
  year={2024},
  eprint={2412.19437},
  archivePrefix={arXiv},
  primaryClass={cs.CL},
  doi={10.48550/arXiv.2412.19437},
  url={https://arxiv.org/abs/2412.19437}
}

@misc{deepseek2025v31,
  title={{DeepSeek-V3.1} Release},
  author={{DeepSeek-AI}},
  year={2025},
  month=aug,
  howpublished={\url{https://api-docs.deepseek.com/news/news250821/}}
}

@misc{anthropic2025claude45,
  title={Introducing {Claude Haiku 4.5}},
  author={{Anthropic}},
  year={2025},
  month=oct,
  howpublished={\url{https://www.anthropic.com/news/claude-haiku-4-5}}
}

@misc{anthropic2025claudesonnet45,
  title={Introducing {Claude Sonnet 4.5}},
  author={{Anthropic}},
  year={2025},
  month=sep,
  howpublished={\url{https://www.anthropic.com/news/claude-sonnet-4-5}}
}

@misc{anthropic2026claudesonnet46,
  title={Introducing {Claude Sonnet 4.6}},
  author={{Anthropic}},
  year={2026},
  month=feb,
  howpublished={\url{https://www.anthropic.com/news/claude-sonnet-4-6}}
}

@inproceedings{shinn2023reflexion,
  title={Reflexion: Language Agents with Verbal Reinforcement Learning},
  author={Shinn, Noah and Cassano, Federico and Gopinath, Ashwin and Narasimhan, Karthik and Yao, Shunyu},
  booktitle={Advances in Neural Information Processing Systems},
  volume={36},
  pages={8634--8652},
  year={2023},
  doi={10.52202/075280-0377},
  url={https://proceedings.neurips.cc/paper_files/paper/2023/hash/1b44b878bb782e6954cd888628510e90-Abstract-Conference.html}
}

@inproceedings{khattab2023dspycompilingdeclarativelanguage,
  title={{DSPy}: Compiling Declarative Language Model Calls into State-of-the-Art Pipelines},
  author={Khattab, Omar and Singhvi, Arnav and Maheshwari, Paridhi and Zhang, Zhiyuan and Santhanam, Keshav and {Vardhamanan A}, Sri and Haq, Saiful and Sharma, Ashutosh and Joshi, Thomas T. and Moazam, Hanna and Miller, Heather and Zaharia, Matei and Potts, Christopher},
  booktitle={The Twelfth International Conference on Learning Representations},
  year={2024},
  url={https://openreview.net/forum?id=sY5N0zY5Od}
}

@misc{tarski:github:18,
  author = {Guillem Franc\'{e}s and Miquel Ramirez and Collaborators},
  title = {Tarski: An {AI} Planning Modeling Framework},
  year = {2018},
  publisher = {{GitHub}},
  journal = {{GitHub} repository},
  howpublished = {\url{https://github.com/aig-upf/tarski}}
}

@INPROCEEDINGS{1374201,
  author={Howey, R. and Long, D. and Fox, M.},
  booktitle={16th IEEE International Conference on Tools with Artificial Intelligence},
  title={{VAL}: Automatic Plan Validation, Continuous Effects and Mixed Initiative Planning Using {PDDL}},
  year={2004},
  pages={294--301},
  publisher={IEEE Computer Society},
  doi={10.1109/ICTAI.2004.120},
  url={https://doi.org/10.1109/ICTAI.2004.120}
}

@article{slaney2001blocks,
  title={Blocks World revisited},
  author={Slaney, John and Thi{\'e}baux, Sylvie},
  journal={Artificial Intelligence},
  volume={125},
  number={1--2},
  pages={119--153},
  year={2001},
  publisher={Elsevier},
  doi={10.1016/S0004-3702(00)00079-5},
  url={https://doi.org/10.1016/S0004-3702(00)00079-5}
}

@misc{seipp-et-al-zenodo2022,
  author={Seipp, Jendrik and Torralba, {\'A}lvaro and Hoffmann, J{\"o}rg},
  title={{PDDL} Generators},
  publisher={Zenodo},
  year={2022},
  doi={10.5281/zenodo.6382173},
  howpublished={\url{https://doi.org/10.5281/zenodo.6382173}}
}
}

\clearpage


\appendix
\clearpage
\section{Analysis: The Valid-to-Success Gap}
\label{app:valid_success_gap}

While L-ICL dramatically improves constraint adherence, a gap often remains between validity and success. This gap is most pronounced in Full Sokoban, where L-ICL achieves 46\% valid plans but only 20\% success (Table~\ref{tab:hero}). Understanding this gap illuminates both L-ICL's strengths and its limitations.

\subsection{Trap Rate Analysis}
\label{app:trap_rate}

In Sokoban, certain states are \emph{traps}: configurations from which the goal is unreachable regardless of future actions (e.g., a box pushed into a corner). We measure the \emph{adjusted trap rate}: among valid plans, what fraction enters a trap state?

\begin{figure}[t]
  \centering
  \includegraphics[width=\columnwidth]{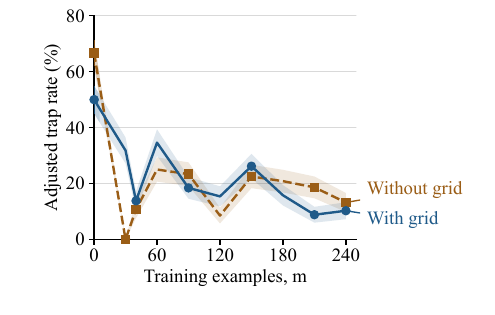}
  \caption{Trap rate decreases with L-ICL. Adjusted trap rate (fraction of valid plans entering unsolvable states) on Sokoban Grid. L-ICL reduces trap rates from 50\% to 10\%, indicating partial learning of strategic constraints.}
  \label{fig:trap_rate}
\end{figure}

Figure~\ref{fig:trap_rate} shows that L-ICL \emph{reduces} trap rates. On Sokoban Grid, the adjusted trap rate drops from 50\% at $m{=}0$ to 10\% at $m{=}210$. This indicates that L-ICL teaches not only immediate constraint satisfaction but also some degree of trap avoidance.

However, the absolute trap rate remains non-negligible, and the valid-to-success gap persists. We hypothesize that trap avoidance requires multi-step lookahead that localized corrections cannot fully provide. A correction like ``pushing box B east from (3,4) is valid'' does not encode that this push leads to an unsolvable configuration three moves later. Addressing this limitation may require complementary approaches such as search or learned value functions.

\subsection{Multi-Object State Tracking}

Comparing Sokoban Grid (no boxes) to Full Sokoban reveals the cost of multi-object tracking. With identical spatial layouts, Sokoban Grid achieves 49\% success while Full Sokoban reaches only 20\%. The difference lies in state complexity: Full Sokoban requires tracking the agent position \emph{and} all box positions, with constraints that depend on their joint configuration.

This difficulty is also evident in BlocksWorld, where \emph{every} object is dynamic. L-ICL improves BlocksWorld success from 48\% to 66\%, but a gap remains between validity (68\%) and success. The pattern suggests that relational constraint learning, while improved by L-ICL, remains more challenging than spatial constraint learning.

\subsection{Decomposing Planning Difficulty}

The valid-to-success gap reveals a clean decomposition of planning difficulty:

\begin{enumerate}
    \item \textbf{Constraint satisfaction}: Generating actions that respect domain physics. L-ICL addresses this effectively across all domains.
    \item \textbf{Strategic selection}: Among valid actions, choosing those that lead toward the goal without entering traps. This requires multi-step reasoning that localized corrections do not directly provide.
\end{enumerate}

This decomposition suggests a practical architecture: use L-ICL to harden constraint satisfaction, then layer strategic reasoning (search, learned policies, or hierarchical planning) on top. The hardened base ensures that any action proposed by the strategic layer is physically valid, separating concerns and simplifying both components.

Table~\ref{tab:valid_success_breakdown} summarizes the valid-to-success gaps across domains, highlighting where strategic failures dominate.

\begin{table}[t]
\centering
{
\small
\setlength{\tabcolsep}{1mm}
\begin{tabular}{@{}lccc@{}}
\toprule
\textbf{Domain} & \textbf{\% Valid} & \textbf{\% Success} & \textbf{Gap} \\
\midrule
8$\times$8 Grid & 89 & 89 & 0 \\
10$\times$10 Maze & 57 & 27 & 30 \\
Sokoban Grid & 63 & 49 & 14 \\
Full Sokoban & 46 & 20 & 26 \\
BlocksWorld & 68 & 66 & 2 \\
\bottomrule
\end{tabular}
}
\caption{Valid-to-success gap analysis across domains with L-ICL[$m{=}60$]. Larger gaps indicate that constraint satisfaction alone is insufficient---strategic reasoning is the bottleneck.}
\label{tab:valid_success_breakdown}
\end{table}

The 8$\times$8 gridworld shows no gap: once constraints are satisfied, the simple structure makes goal-reaching straightforward. The 10$\times$10 maze and Full Sokoban show the largest gaps, reflecting the strategic complexity of navigating dead ends and avoiding irreversible trap states, respectively. BlocksWorld shows a small gap, suggesting that while relational constraints are harder to learn, once learned they suffice for task completion in our 5-block instances.

\section{Out-of-Distribution Generalization}
\label{app:ood}

\begin{figure}[t]
  \centering
  \begin{subfigure}[b]{0.48\columnwidth}
    \centering
    \includegraphics[width=\textwidth]{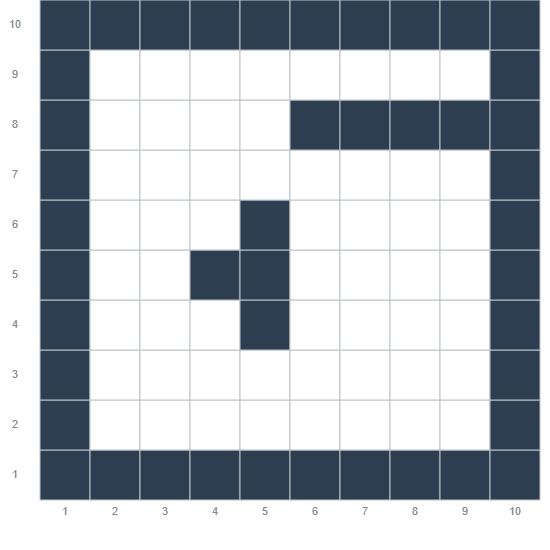}
    \caption{10$\times$10 maze (training distribution)}
    \label{fig:ood_10x10}
  \end{subfigure}
  \hfill
  \begin{subfigure}[b]{0.48\columnwidth}
    \centering
    \includegraphics[width=\textwidth]{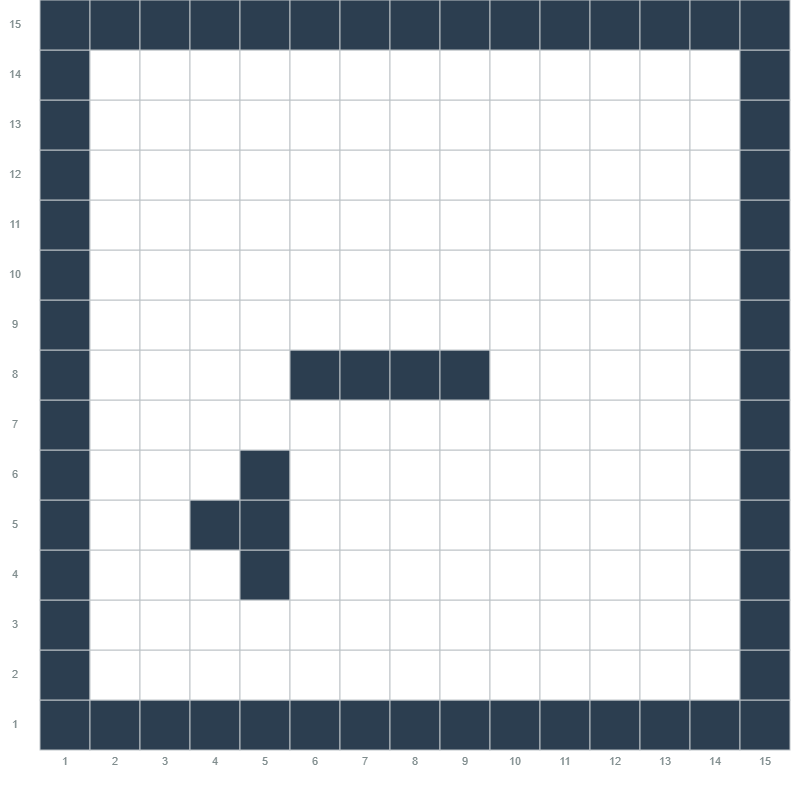}
    \caption{15$\times$15 maze (OOD evaluation)}
    \label{fig:ood_15x15}
  \end{subfigure}
  \caption{Out-of-distribution generalization setup. L-ICL corrections are accumulated on 10$\times$10 mazes (left) and evaluated on 15$\times$15 mazes (right). The larger mazes contain positions not seen during training, yet corrections transfer substantially, despite the penalty for boundary violations differing.}
  \label{fig:ood_domains}
\end{figure}

A key question for any learning-based approach is whether acquired knowledge transfers beyond the training distribution. For L-ICL, this translates to: do corrections learned on smaller problem instances improve performance on larger, unseen instances? We investigate this by training L-ICL on 10$\times$10 mazes and evaluating on 15$\times$15 mazes, showin in Figure~\ref{fig:ood_domains}.

\subsection{Experimental Setup}

We accumulate L-ICL corrections using the standard training procedure on 10$\times$10 maze instances (Section~\ref{sec:setup}). We then evaluate the resulting prompts on a held-out test set of 100 15$\times$15 mazes. The larger mazes are generated using the same procedural algorithm (randomized depth-first search) with proportionally scaled wall density, but contain positions and path structures never seen during training. 

\subsection{Results}

\begin{figure}[t]
  \centering
  \includegraphics[width=\columnwidth]{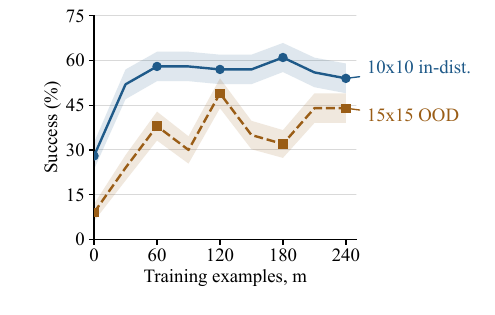}
  \caption{Out-of-distribution transfer: 10$\times$10 $\to$ 15$\times$15. Corrections learned on 10$\times$10 mazes transfer to larger instances, improving success from 9\% to 49\%. A gap remains compared to in-distribution performance (57\% on 10$\times$10), but transfer is substantial.}
  \label{fig:ood}
\end{figure}

Figure~\ref{fig:ood} shows that L-ICL corrections provide substantial transfer to larger instances. At $m{=}0$ (no corrections), the 15$\times$15 maze achieves only 9\% success---comparable to the 10$\times$10 baseline without corrections. With corrections accumulated from 10$\times$10 training instances, 15$\times$15 success improves to 49\% at $m{=}120$, representing a \textbf{5$\times$ improvement} over the no-correction baseline.

Table~\ref{tab:ood_summary} summarizes the transfer results at key checkpoints.

\begin{table}[t]
\centering
{
\small
\setlength{\tabcolsep}{1mm}
\begin{tabular}{@{}lcc@{}}
\toprule
\textbf{Training Examples} & \textbf{10$\times$10 (in-dist.)} & \textbf{15$\times$15 (OOD)} \\
\midrule
$m = 0$ & 16 & 9 \\
$m = 30$ & 21 & 18 \\
$m = 60$ & 27 & 31 \\
$m = 120$ & 57 & 49 \\
\bottomrule
\end{tabular}
}
\caption{Out-of-distribution generalization: corrections trained on 10$\times$10 mazes evaluated on 15$\times$15 mazes. We report success rate (\%) and compare to in-distribution 10$\times$10 performance.}
\label{tab:ood_summary}
\end{table}

\subsection{Why Does Transfer Work?}

The transfer is notable because 15$\times$15 mazes contain positions (e.g., $(12, 14)$) and wall configurations that never appear in 10$\times$10 training instances. We hypothesize that corrections transfer because they encode \emph{constraint types} rather than \emph{specific positions}. 

Consider a correction like:
\begin{lstlisting}
>>> get_applicable_actions((3, 4))
['move_north', 'move_south']
\end{lstlisting}

While this example specifies position $(3, 4)$, it implicitly teaches a general principle: when east and west are blocked (by walls or boundaries), only north and south are valid. The LLM can generalize this pattern to novel positions in larger grids.

This interpretation is supported by the observation that transfer improves with more corrections ($m$). Early corrections address common constraint patterns (boundary violations, simple wall configurations); as $m$ increases, rarer patterns are covered, and the accumulated examples provide a richer specification that generalizes more robustly.

\subsection{Transfer Gap Analysis}

While transfer is substantial, a gap remains between in-distribution and OOD performance (57\% vs. 49\% at $m{=}120$). We identify two contributing factors:

\begin{enumerate}
    \item \textbf{Unseen spatial configurations}: Larger mazes contain junction types and corridor patterns that may not appear in smaller instances. Some constraint violations specific to these configurations are not addressed by 10$\times$10 training.
    
    \item \textbf{Longer planning horizons}: 15$\times$15 mazes require longer plans, providing more opportunities for errors to accumulate. Even with improved per-step validity, the probability of completing an error-free trajectory decreases with plan length.
\end{enumerate}

These findings suggest that for maximum OOD performance, practitioners should either (a) train on a mixture of problem sizes, or (b) accept a modest performance gap when deploying to larger instances than those seen during training.

\subsection{Cross-Domain Transfer}

We also conducted preliminary experiments on cross-domain transfer: using corrections from one domain (e.g., 8$\times$8 gridworld) to improve another (e.g., 10$\times$10 maze). Results were mixed---corrections for basic movement constraints (boundary checking) transferred, but domain-specific spatial structures (two-room vs. maze corridors) did not. This suggests that L-ICL learns a combination of general procedural knowledge and domain-specific constraint instantiations, with only the former transferring across domains.

 \section{Additional Mechanism and Robustness Ablations}
  \label{app:mechanism_ablations}

  All tables in this section use the metric definitions from
  Section~\ref{sec:metrics}: $V$ is the percentage of generated plans
  containing no inapplicable action, and $S$ is the percentage that both
  contain no inapplicable action and reach the goal. Each condition
  contains 100 held-out 10$\times$10 Maze problems.

  \subsection{Correction-Selection Strategy}
  \label{app:correction_strategy}

  We compare four ways of selecting correction states. First-error
  localization retains the first observed error of each enabled
  subroutine. Last-error localization retains the final such error,
  all-error localization retains all observed errors, and the random
  control inserts a correct subroutine example from a randomly selected
  state rather than an observed failure. All conditions use Claude Haiku
  4.5 and 60 training problems.

  \begin{table}[t]
  \centering
  \small
  \setlength{\tabcolsep}{1.5mm}
  \begin{tabular}{@{}lcccc@{}}
  \toprule
  & \multicolumn{2}{c}{\textbf{With Grid}}
  & \multicolumn{2}{c}{\textbf{Without Grid}} \\
  \textbf{Selection strategy} & V & S & V & S \\
  \midrule
  First error            & \textbf{59} & \textbf{57}
                         & 32 & \textbf{30} \\
  All errors             & 56 & 56 & \textbf{36} & 28 \\
  Last error             & 48 & 46 & 24 & 18 \\
  Random correct example & 52 & 51 & 8 & 8 \\
  \bottomrule
  \end{tabular}
  \caption{Correction-selection ablation using Claude Haiku 4.5 at
  $m{=}60$. First-error localization has the highest success in both
  representation conditions and the highest validity with the grid.}
  \label{tab:correction_strategy_ablation}
  \end{table}

  The comparison does not show that first-error localization is strictly
  best on every metric: all-error localization is four points higher in
  no-grid validity. It does show that first-error localization provides
  the strongest overall tradeoff while avoiding corrections derived from
  later states. It also shows that failure-targeted examples are
  substantially more useful than randomly selected correct examples when
  no visualization is available.

  \subsection{Natural-Language Corrections Without PTP}
  \label{app:cot_corrections}

  To separate the correction mechanism from the PTP representation, we
  apply first-error accumulation to standard Chain-of-Thought prompts.
  No partial program or doctest syntax is used. Instead, each correction
  is rendered as a natural-language planning note and appended to the
  prompt. These experiments use Claude Haiku 4.5 and 60 training
  problems.

  \begin{table}[t]
    \centering
    \small
    \setlength{\tabcolsep}{1.5mm}
    \begin{tabular}{@{}lcccc@{}}
    \toprule
    & \multicolumn{2}{c}{\textbf{With Grid}}
    & \multicolumn{2}{c}{\textbf{Without Grid}} \\
    \textbf{Prompt} & V & S & V & S \\
    \midrule
    CoT, no corrections
        & 26 & 19 & 4 & 0 \\
    CoT + NL corrections
        & \textbf{63} & \textbf{59}
        & \textbf{55} & \textbf{53} \\
    \bottomrule
    \end{tabular}
    \caption{Natural-language correction ablation with Claude Haiku 4.5
    at $m{=}60$, evaluated on $n{=}100$ held-out problems per condition.
    Corrections are deterministic natural-language renderings
    of first-invalid-action feedback produced by the symbolic evaluator.
    $V$ denotes strict action-sequence validity and $S$ denotes successful
    goal completion. Natural-language corrections substantially improve both
    metrics with and without the grid representation.}
    \label{tab:cot_correction_ablation}
  \end{table}

  The result establishes that failure-driven correction accumulation, i.e. L-ICL, is
  not specific to PTP. PTP remains useful because its named subroutines
  provide stable, automatically addressable locations for compact
  corrections.  Furthermore, these results indicate that L-ICL works regardless of format-
ting, with the main advantage of PTP over natural language
being the ability to guarantee L-ICL insertability through
parsed corrections.

  \subsection{Imperfect Correction Feedback With and Without the Grid}
  \label{app:imperfect_feedback_full}

  We test robustness in two ways. First, we independently corrupt each
  symbolic-oracle output used to render a correction with probability
  $\eta$. A corrupted action set is produced by randomly adding an
  incorrect action, removing a correct action, or replacing one action
  with another. Second, we use Claude Sonnet 4.6 to generate the
  correction output. 
  All conditions use Claude Haiku 4.5 as the planning model, $m{=}60$
  training problems, and 100 held-out test problems.

  \subsection{Frozen-Prompt Transfer Across Model Families}
  \label{app:frozen_prompt_transfer}

  \begin{table}[t]
  \centering
  \small
  \setlength{\tabcolsep}{2mm}
  \begin{tabular}{@{}lcc@{}}
  \toprule
  \textbf{Evaluation model} &
  \textbf{With Grid V} &
  \textbf{Without Grid V} \\
  \midrule
  DeepSeek V3       & 40 & \textbf{57} \\
  Claude Sonnet 4.5 & \textbf{90} & 52 \\
  Claude Haiku 4.5  & 68 & 29 \\
  \bottomrule
  \end{tabular}
  \caption{Applicability of plans generated from the same frozen
  DeepSeek-V3 correction prompt. The corrections can be consumed by
  other model families without model-specific regeneration.}
  \label{tab:frozen_prompt_transfer}
  \end{table}
  We accumulate a correction prompt from 60 DeepSeek V3 training
  problems, freeze it, and evaluate the identical prompt using DeepSeek
  V3, Claude Sonnet 4.5, and Claude Haiku 4.5 in Table \ref{tab:frozen_prompt_transfer}. No recipient-model
  training or correction regeneration is performed.
  This experiment establishes prompt compatibility across model
  families. It is not, by itself, a within-model estimate of the causal
  gain from corrections because the recipient-model rows do not include
  corresponding empty-correction controls. The model-specific learning
  curves in Figure~\ref{fig:llm_ablation} provide that complementary
  comparison.

\section{Domain Specifications}
\label{app:domains}

This appendix provides detailed specifications of the experimental domains used in our evaluation. For each domain, we describe the state representation, action space, constraints and goal conditions.
\begin{table}[t]
\centering
{
\small
\setlength{\tabcolsep}{1mm}
\begin{tabular}{@{}lcccc@{}}
\toprule
\textbf{Domain} & \textbf{Grid} & \textbf{Actions} & \textbf{Objects} & \textbf{Irreversible} \\
\midrule
8$\times$8 Grid & Simple & 4 & 1 & No \\
10$\times$10 Maze & Complex & 4 & 1  & No \\
Sokoban Grid & Complex & 4 & 1  & No \\
Full Sokoban & Complex & 8 & 3  & Yes \\
BlocksWorld & None & 3 & 5 & No \\
\bottomrule
\end{tabular}
}
\caption{Progressive ablation across experimental domains. Each domain adds complexity along one or more axes while controlling others.}
\label{tab:domain_progression}
\end{table}

\subsection{8$\times$8 Two-Room Gridworld}
\label{app:domain-gridworld}

\paragraph{State Space.} The state consists of the agent's $(x, y)$ position on an 8$\times$8 grid. Coordinates range from $(1, 1)$ at the bottom-left to $(8, 8)$ at the top-right.

\paragraph{Environment Structure.} The grid is divided into two rooms by a vertical wall running through column 5, with a single doorway allowing passage between rooms (doorway position varies by instance). Start positions are randomly sampled from one room, and goal positions from the other, ensuring all paths must traverse the doorway.

\paragraph{Action Space.} Four actions: \texttt{move\_north} $(+y)$, \texttt{move\_south} $(-y)$, \texttt{move\_east} $(+x)$, and \texttt{move\_west} $(-x)$.

\paragraph{Constraints.} An action is \emph{valid} if and only if:
\begin{enumerate}
    \item The resulting position remains within grid bounds.
    \item The movement does not cross a wall segment.
\end{enumerate}

\paragraph{Goal Condition.} The agent's position equals the goal position.

\paragraph{Optimal Solution.} The shortest path between start and goal, computed via breadth-first search. Optimal paths typically require 8--12 steps.

\begin{figure}[t]
    \centering
 \includegraphics[width=\columnwidth]{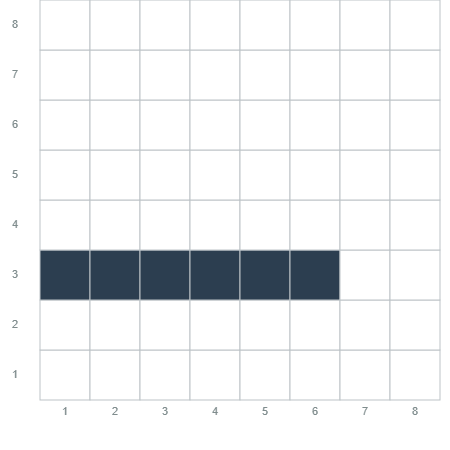}
    \caption{Example 8$\times$8 two-room gridworld instance. Walls are shown as filled cells.}
    \label{fig:gridworld_example}
\end{figure}

\subsection{10$\times$10 Maze}
\label{app:domain-maze}

\paragraph{State Space.} The state consists of the agent's $(x, y)$ position on a 10$\times$10 grid. Coordinates range from $(1, 1)$ to $(10, 10)$.

\paragraph{Environment Structure.} Mazes are procedurally generated using a randomized depth-first search algorithm, producing a spanning tree of corridors with exactly one path between any two open cells. This ensures unique shortest paths and creates narrow corridors with dead ends that require backtracking if the agent makes suboptimal choices.

\paragraph{Action Space.} Four actions: \texttt{move\_north}, \texttt{move\_south}, \texttt{move\_east}, \texttt{move\_west}.

\paragraph{Constraints.} Identical to the 8$\times$8 gridworld: actions must keep the agent in bounds and cannot cross walls.

\paragraph{Goal Condition.} The agent's position equals the goal position.

\paragraph{Optimal Solution.} The unique shortest path through the maze.

\begin{figure}[t]
    \centering
    \includegraphics[width=\columnwidth]{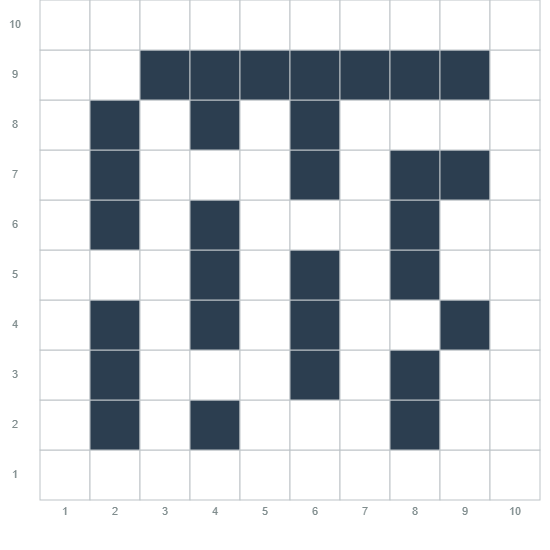}
    \caption{Example 10$\times$10 maze instance. The maze structure creates narrow corridors and dead ends, requiring longer plans than the two-room gridworld.}
    \label{fig:maze_example}
\end{figure}

\subsection{Sokoban-Style Gridworld}
\label{app:domain-sokoban-grid}

\paragraph{State Space.} The state consists of the agent's $(x, y)$ position on a grid that uses Sokoban-style layouts. Coordinates are 1-indexed.

\paragraph{Environment Structure.} We use grid layouts from standard Sokoban benchmarks but remove all pushable boxes. The layouts retain walls, open floor cells, and the spatial structure of Sokoban puzzles, including irregular room shapes and narrow passages. This domain serves as an ablation to isolate the effect of Sokoban's spatial complexity from the challenge of multi-object state tracking.

\paragraph{Action Space.} Four actions: \texttt{move\_north}, \texttt{move\_south}, \texttt{move\_east}, \texttt{move\_west}.

\paragraph{Constraints.} Actions must keep the agent within the walkable floor area and cannot cross walls.

\paragraph{Goal Condition.} The agent reaches a designated goal cell.

\begin{figure}[t]
    \centering
    \includegraphics[width=\columnwidth]{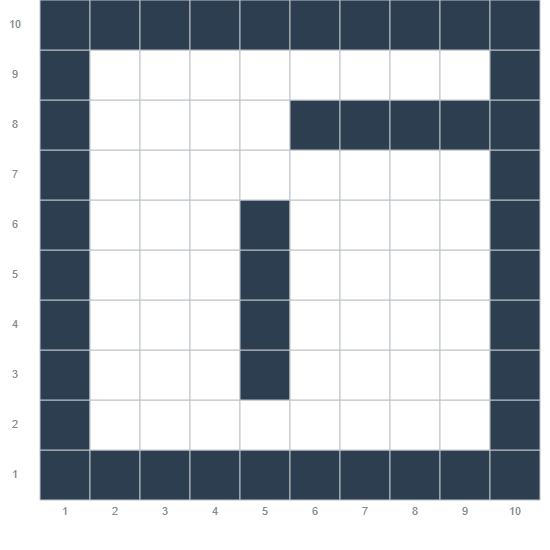}
    \caption{Example Sokoban-style gridworld instance. The layout is derived from a Sokoban puzzle but contains no pushable boxes, isolating spatial navigation from object manipulation.}
    \label{fig:sokoban_grid_example}
\end{figure}

\subsection{Full Sokoban}
\label{app:domain-sokoban}

\paragraph{State Space.} The state consists of:
\begin{itemize}
    \item The agent's $(x, y)$ position (1-indexed).
    \item The box position $(x, y)$.
\end{itemize}
Our instances use 1 box.

\paragraph{Environment Structure.} Standard Sokoban puzzle layouts from established benchmarks, including walls, floor cells, and designated target locations where boxes must be placed.

\paragraph{Action Space.} Eight actions:
\begin{itemize}
    \item Movement: \texttt{move\_north}, \texttt{move\_south}, \texttt{move\_east}, \texttt{move\_west}---move the agent one cell in the specified direction if the destination is empty floor.
    \item Pushing: \texttt{push\_north}, \texttt{push\_south}, \texttt{push\_east}, \texttt{push\_west}---move the agent into a cell containing a box, pushing the box one cell further in the same direction.
\end{itemize}

\paragraph{Constraints.} An action is \emph{valid} if and only if:
\begin{enumerate}
    \item \textbf{Movement}: The destination cell is within bounds, is not a wall, and does not contain a box.
    \item \textbf{Pushing}: The cell adjacent to the agent contains a box, and the cell beyond the box (in the push direction) is within bounds, is not a wall, and does not contain another box.
\end{enumerate}

\paragraph{Irreversibility.} Unlike navigation domains, Sokoban contains \emph{trap states}---configurations from which the goal is unreachable. Common traps include:
\begin{itemize}
    \item Pushing a box into a corner (cannot be retrieved).
    \item Pushing a box against a wall such that it cannot reach any target.
\end{itemize}

\paragraph{Goal Condition.} The box occupies the shown target position.

\begin{figure}[t]
    \centering
    \includegraphics[width=\columnwidth]{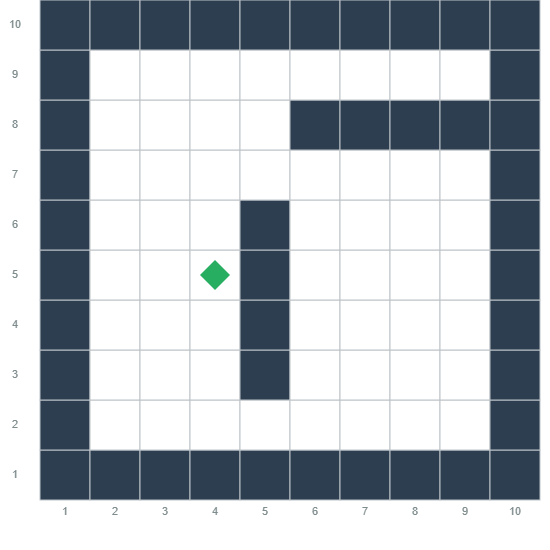}
    \caption{Example Sokoban instance. The agent must push the box onto the target location without creating deadlocks.}
    \label{fig:sokoban_example}
\end{figure}

\subsection{BlocksWorld}
\label{app:domain-blocksworld}

\paragraph{State Space.} The state consists of a configuration of $n$ uniquely labeled blocks (we use $n=5$ in our experiments). Each block is either:
\begin{itemize}
    \item On the table, or
    \item On top of exactly one other block.
\end{itemize}
A block is \emph{clear} if no other block is on top of it. The table has unlimited capacity.

\paragraph{Action Space.} Three actions, described in natural language:

\begin{enumerate}
    \item \textbf{Move block from block to block} (\texttt{move-b-to-b}): Pick up a block that is currently sitting on top of another block and place it onto a third block. This requires that the block being moved has nothing on top of it (is clear) and that the destination block also has nothing on top of it (is clear). After the move, the block that was underneath the moved block becomes clear.
    
    \item \textbf{Move block from block to table} (\texttt{move-b-to-t}): Pick up a block that is currently sitting on top of another block and place it on the table. This requires that the block being moved has nothing on top of it (is clear). After the move, the block that was underneath becomes clear, and the moved block is now on the table.
    
    \item \textbf{Move block from table to block} (\texttt{move-t-to-b}): Pick up a block that is currently on the table and place it onto another block. This requires that both the block being moved and the destination block have nothing on top of them (are clear). After the move, the destination block is no longer clear.
\end{enumerate}

\paragraph{Constraints.} The preconditions for each action are:
\begin{itemize}
    \item \texttt{move-b-to-b($b_m$, $b_f$, $b_t$)}: Block $b_m$ is clear, block $b_t$ is clear, $b_m$ is currently on $b_f$, and $b_m \neq b_t$.
    \item \texttt{move-b-to-t($b_m$, $b_f$)}: Block $b_m$ is clear and $b_m$ is currently on $b_f$.
    \item \texttt{move-t-to-b($b_m$, $b_t$)}: Block $b_m$ is clear, block $b_t$ is clear, $b_m$ is currently on the table, and $b_m \neq b_t$.
\end{itemize}

\paragraph{Goal Condition.} The block configuration matches a target specification, typically given as a set of \texttt{on($b_1$, $b_2$)} predicates describing which blocks must be stacked on which.

\paragraph{Differences from Navigation Domains.} BlocksWorld differs qualitatively from the grid-based domains:
\begin{itemize}
    \item \textbf{No spatial structure}: Constraints are purely relational (``block A is on block B'') rather than geometric.
    \item \textbf{All objects are dynamic}: Every block can be moved, unlike navigation where only the agent moves.
    \item \textbf{Algorithmic solutions}: We additionally provide an algorithmic sketch (the Universal Blocksworld Algorithm~\citep{stechly2024chain}) to test whether L-ICL can improve adherence to prescribed planning strategies.
\end{itemize}


\section{Baseline Method Implementations}
\label{app:baselines}

This appendix provides detailed specifications of all baseline methods evaluated in our experiments. All baselines operate on the same task: given a problem description with start position, goal position, walls, and (optionally) deadzones, produce an action sequence to navigate from start to goal. We organize baselines into two categories: \emph{prompt-only} methods that rely solely on LLM reasoning, and \emph{oracle} methods that receive feedback from a ground-truth simulator.

\subsection{Prompt-Only Baselines}

\subsubsection{Zero-Shot Chain-of-Thought (Zero-Shot CoT)}

The simplest baseline provides the model with task instructions, and asks it to reason step-by-step to produce a navigation plan.

\paragraph{Implementation.} The prompt includes: (1) a task description explaining gridworld navigation, valid actions, and movement constraints; (2) an ASCII representation of the problem (if applicable); (3) the query problem with start/goal coordinates; and (4) output format instructions requiring \texttt{**Final Action Sequence:** action1, action2, ...}. We use temperature 1.0 for all experiments unless otherwise noted.

\subsubsection{RAG-CoT (Retrieval-Augmented Chain-of-Thought)}

This baseline extends Zero-Shot CoT with dynamic example selection based on similarity to the query problem. We retrieve the most relevant training examples within a character budget (10,000 or 20,000 characters in our experiments).

\paragraph{Similarity Metric.} We compute similarity based on Manhattan distance between start-goal pairs:
{\scriptsize
\begin{equation}
\text{similarity}(q, c) = \frac{1}{1 + |d_q - d_c|}
\end{equation}
}
where $d = |g_x - s_x| + |g_y - s_y|$ is the Manhattan distance from start $s$ to goal $g$. This metric prefers training examples with similar navigation distances, under the assumption that problems with similar start-to-goal distances share structural similarities.

\paragraph{Retrieval Modes.} We evaluate three retrieval strategies:
\begin{itemize}
    \item \textbf{Strict}: Add examples until the budget would be exceeded (conservative).
    \item \textbf{Generous}: Add examples until the budget is just crossed (permissive).
    \item \textbf{Fixed}: Include fixed examples plus retrieved examples up to the remaining budget.
\end{itemize}

Our main experiments use the generous mode.

\subsubsection{Self-Consistency}

Self-Consistency~\citep{wang2023selfconsistency} generates multiple independent reasoning trajectories and selects the final answer via majority voting.

\paragraph{Implementation.} We sample $k=5$ independent CoT traces using temperature 1.0 for diversity. Each sample uses the same prompt with an annotation indicating the sample number (e.g., ``Sample 3/5''). We parse action sequences from each sample, count votes for each unique plan (exact sequence match), and select the plan with the highest vote count. For tie-breaking, we use an additional LLM call to evaluate candidates based on their self-critique annotations.

\paragraph{Self-Critique.} Each sample includes a self-critique section where the model evaluates its own reasoning, providing confidence estimates and noting potential issues. This information is used only for tie-breaking.

\subsubsection{Self-Refine}

Self-Refine~\citep{madaan2023selfrefine} allows the model to iteratively review and improve its own solutions without external feedback.

\paragraph{Implementation.} The model generates an initial attempt, then receives up to $N=5$ refinement opportunities. In each refinement round, the model sees its previous response and is instructed to check for potential mistakes: boundary violations, wall collisions, goal reachability, deadzone avoidance, and path optimality. The model may either provide a corrected plan or explicitly state ``\texttt{**No further refinement needed.**}''

\paragraph{Termination Conditions.} Refinement stops when: (1) the model explicitly states satisfaction or (2) maximum refinement steps are reached

\paragraph{Key Distinction.} Unlike oracle baselines, Self-Refine receives \emph{no} external feedback about plan validity. The model must introspect on its own reasoning, which prior work has shown to be unreliable for planning tasks~\citep{stechly2025on}.

\subsubsection{ReAct (Prompt-Only)}

ReAct~\citep{yao2023react} interleaves reasoning and action selection in a textual trace format.

\paragraph{Implementation.} The model alternates between \texttt{Thought:} steps (reasoning about current state and next move) and \texttt{Action:} steps (single movement action). All reasoning and actions are generated in a single LLM call---no external tool execution occurs. The prompt includes guidelines to keep moves consistent with the grid layout, avoid illegal steps, provide reasoning before each action, and end with an explicit final action sequence.

\paragraph{Trace Format.}
\begin{lstlisting}
Thought: [analyze current state and next move]
Action: move-direction
Thought: [continue reasoning]
Action: move-direction
...
Final Thought: [summarize path to goal]
**Final Action Sequence:** action1, action2, ...
\end{lstlisting}

\subsubsection{Tree-of-Thoughts (Prompt-Only)}

Tree-of-Thoughts (ToT)~\citep{yao2023tree} explores multiple reasoning paths through iterative expansion and scoring.

\paragraph{Implementation.} We use a prompt-only variant with breadth-first tree search. Parameters: breadth $b=5$ (nodes per level), depth $d=3$ (expansion rounds), max step actions $m=8$ (actions per candidate). At each depth level, the model generates candidate continuations as JSON, including a thought description, proposed actions, confidence score (0--100), terminal flag, and optional final plan. We keep the top-$k$ nodes by confidence (beam search) and finalize by selecting the best terminal node or completing the top non-terminal node.

\paragraph{Scoring.} Without oracle access, scoring uses only LLM self-assessed confidence and plan length:
{\scriptsize
\begin{equation}
\text{score} = (\text{confidence}, -\text{plan\_length})
\end{equation}
}
Higher confidence is preferred; shorter plans are preferred as a tiebreaker.

\subsection{Oracle Baselines}

Oracle baselines have access to a ground-truth environment simulator that provides feedback on plan validity. This represents an upper bound on what prompt-only methods could achieve with perfect self-verification.

\subsubsection{ReAct (+Oracle f/b)}

This two-step approach allows the model to receive targeted feedback about errors and produce a corrected plan.

\paragraph{Implementation.} Step 1: Generate an initial CoT plan. Step 2: If errors are detected by the oracle, provide specific feedback and request correction. Maximum 2 LLM calls per problem. We use temperature 0.3 for more deterministic outputs in the correction step.

\paragraph{Feedback Types.} The oracle provides two types of feedback:
\begin{enumerate}
    \item \textbf{Invalid Move}: ``Your plan has an ERROR at step $N$. The action `move-X' at position $(x,y)$ is INVALID because it would move into a wall or out of bounds.''
    \item \textbf{Incomplete Path}: ``Your plan is INCOMPLETE. After executing all $N$ actions, you ended at position $(x,y)$ but did not reach the goal.''
\end{enumerate}

\paragraph{Key Distinction.} ReAct (+Oracle f/b) queries the verifier at \emph{test time} for each proposed plan, while L-ICL uses the oracle \emph{only during training}. At inference, L-ICL requires a single forward pass with no external dependencies.

\subsection{Evaluation Infrastructure}

\paragraph{Action Parsing.} All baselines use the same action sequence parser that handles multiple output formats. We search for explicit patterns (e.g., \texttt{**Final Action Sequence:**}), fall back to lines with comma-separated actions, and as a last resort extract all \texttt{move-*} actions from the response. Actions are normalized to canonical form (e.g., ``north'' $\to$ ``move-north'').

\paragraph{Plan Validation.} Plan validity is determined by simulating the action sequence:
\begin{enumerate}
    \item Parse problem to extract start/goal positions and obstacles.
    \item Execute actions sequentially, checking bounds and wall collisions.
    \item Verify goal is reached.
    \item Compare plan length to BFS optimal.
\end{enumerate}

\subsection{Summary of Baseline Characteristics}

Table~\ref{tab:baseline_summary} summarizes the key characteristics distinguishing each baseline.

\begin{table}[t]
\centering
{
\small
\setlength{\tabcolsep}{1mm}
\begin{tabular}{@{}lccc@{}}
\toprule
\textbf{Method} & \textbf{Examples} & \textbf{LLM Calls} & \textbf{Tools} \\
\midrule
Zero-Shot CoT & 0 & 1 & none \\
RAG-CoT & retrieved & 1 & none \\
Self-Consistency & fixed (3) & $k$ & none \\
Self-Refine & fixed (3) & $1$--$N$ & none \\
ReAct (Prompt) & fixed (3) & 1 & none \\
ToT (Prompt) & fixed (3) & $O(b \cdot d)$ & none \\
\midrule
ReAct (+Oracle f/b) & fixed (3) & 1--2 & test time \\
\midrule
\textbf{L-ICL (ours)} & localized & 1 & train only \\
\bottomrule
\end{tabular}
}
\caption{Summary of baseline characteristics. ``Examples'' indicates whether the method uses in-context examples; ``LLM Calls'' indicates calls per problem; ``Tools'' indicates whether external tools are used.}
\label{tab:baseline_summary}
\end{table}

\section{Implementation Details}
\label{app:subroutines}

This appendix provides detailed implementation specifications for L-ICL, including the system architecture, correction generation process, and evaluation pipeline. We provide sufficient detail for reproducing our experiments. 

\subsection{System Architecture}
\label{app:architecture}

Our implementation consists of four main components that work together to execute the L-ICL training loop:

\begin{enumerate}
    \item \textbf{Partial Program Generator}: Constructs PTP-style prompts with subroutine documentation and accumulated corrections formatted as input-output examples.
    
    \item \textbf{LLM Interface}: Sends prompts to language models and parses structured traces from responses.
    
    \item \textbf{Evaluation Engine}: Validates generated plans using external tools and step-by-step simulation, identifying the first point of failure.
    
    \item \textbf{Correction Accumulator}: Extracts corrections from evaluation mismatches and injects them into subsequent prompts.
\end{enumerate}

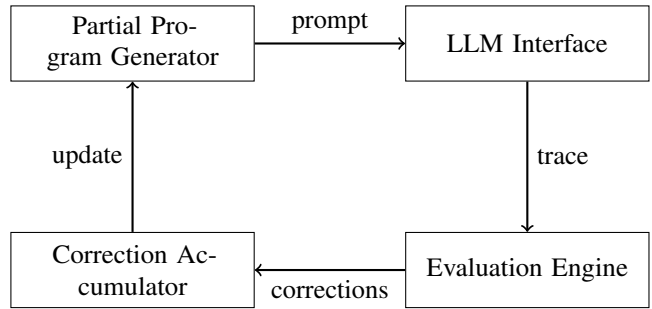
\begin{figure}[t]
    \centering
    \begin{tikzpicture}[node distance=1cm, auto,
        block/.style={rectangle, draw, text width=2.5cm, text centered, minimum height=0.8cm},
        arrow/.style={->, thick}]
        
        \node[block] (pp) {Partial Program Generator};
        \node[block, right=of pp] (llm) {LLM Interface};
        \node[block, below=of llm] (eval) {Evaluation Engine};
        \node[block, left=of eval] (corr) {Correction Accumulator};
        
        \draw[arrow] (pp) -- node[above] {prompt} (llm);
        \draw[arrow] (llm) -- node[right] {trace} (eval);
        \draw[arrow] (eval) -- node[below] {corrections} (corr);
        \draw[arrow] (corr) -- node[left] {update} (pp);
    \end{tikzpicture}
    \caption{System architecture for L-ICL training. The loop iterates over training problems, accumulating corrections that progressively refine the prompt.}
    \label{fig:system_architecture}
\end{figure}

Figure~\ref{fig:system_architecture} illustrates the data flow between these components during L-ICL training.

\subsection{Subroutine Specifications by Domain}
\label{app:subroutine_specs}

Each domain defines a set of planning primitives that the LLM must ``implement'' through trace generation. We describe the subroutines for each domain, including their signatures, semantics, and the constraints they encode.

\subsubsection{Subroutines}

We use the following subroutines in our experiments:

\paragraph{State Extraction.}
\begin{itemize}
    \item \texttt{extract\_initial\_state(problem) $\to$ State}: Parses the problem description to extract the agent's starting position and environment structure.
    \item \texttt{extract\_goal(problem) $\to$ State}: Parses the goal specification from the problem.
\end{itemize}

\paragraph{Action Generation.}
\begin{itemize}
    \item \texttt{get\_applicable\_actions(state, goal) $\to$ Set[Action]}: Returns the set of actions that can be legally executed from the current state. For navigation, this filters the four cardinal directions to exclude moves that would exit the grid or collide with walls.
    \item \texttt{get\_optimal\_actions(state, goal) $\to$ Set[Action]}: Returns the subset of applicable actions that lie on an optimal path to the goal. This is computed using shortest-path algorithms/ a planner. For BlocksWorld, we replace this with  \texttt{get\_recommended\_actions(state, goal) $\to$ Set[Action]}, which returns the set of actions prescribed by the Universal Blocksworld Algorithm.
\end{itemize}

\paragraph{State Transition and Goal Test.}
\begin{itemize}
    \item \texttt{apply\_action(state, action) $\to$ State}: Returns the state resulting from executing the action. For navigation, this updates the agent's coordinates.
    \item \texttt{at\_goal(state, goal) $\to$ bool}: Returns whether the current state satisfies the goal condition.
\end{itemize}

\subsection{Correction Format and Integration}
\label{app:correction_format}

L-ICL corrections are formatted as doctest-style input-output examples that are injected into subroutine documentation. This format is well-represented in LLM training data, facilitating generalization.

\paragraph{Correction Structure.} Each correction consists of three components:
\begin{enumerate}
    \item \textbf{Function identifier}: Which subroutine the correction applies to.
    \item \textbf{Input}: The arguments that triggered the mismatch.
    \item \textbf{Correct output}: The oracle-provided ground truth.
\end{enumerate}

\paragraph{Example Correction.} Consider an LLM that incorrectly proposes moving east from position $(3,4)$ when a wall blocks that direction. The evaluation detects that \texttt{move\_east} is not in the set of applicable actions. L-ICL generates a correction:

\begin{lstlisting}[basicstyle=\footnotesize\ttfamily]
>>> get_applicable_actions(state=(3,4), walls={(3,5)})
{'move_north', 'move_south', 'move_west'}
\end{lstlisting}

This correction is inserted into the documentation for \texttt{get\_applicable\_actions}, providing an explicit example that eastward movement from $(3,4)$ is invalid.

\paragraph{Correction Channels and Domain Configuration.}
\label{app:correction_configuration}

Applicability corrections are enabled in every domain and constitute the feedback analyzed in our correction-specific results. For Gridworld, Maze, Sokoban Grid, and Full Sokoban, we additionally enable an auxiliary optimal-action channel. On each generated trace, the localizer retains the first applicable action that is not in the oracle-provided optimal-action set, if such an action occurs on the valid prefix. It also retains the first invalid action, if one occurs, and stops localization at that point because subsequent states are undefined. Consequently, a problem contributes at most two corrections in these four domains: one auxiliary optimal-action correction and one applicability correction.

BlocksWorld uses only the first applicability correction. In BlocksWorld, the auxiliary channel would supervise adherence to the supplied Universal BlocksWorld strategy rather than a hard legality constraint. Preliminary runs that combined both signals did not yield reliable learning, so all reported BlocksWorld experiments use the applicability-only configuration. This choice is fixed at the domain level and is not selected separately for individual instances or test results.

\paragraph{Correction Accumulation.} Corrections accumulate across training problems. In the standard symbolic-oracle implementation, correction observations are grouped by the exact pair consisting of the subroutine and its input tuple. Repeated observations may remain in the internal accumulator, but the prompt renderer produces only one doctest example for each unique subroutine-input pair. Its correct output is generated by executing the corresponding oracle proxy. After each training batch, the prompt is rebuilt from the base template using all unique inputs accumulated so far, so earlier corrections persist without repeatedly patching an already modified prompt. The batch size is configurable; the reported experiments use batches of ten problems.

\subsection{Evaluation Pipeline}
\label{app:evaluation_pipeline}

Plan evaluation proceeds through multiple validation stages, each providing increasingly detailed feedback.

\subsubsection{External Plan Validation}

We use the VAL validator~\citep{1374201}, the standard tool for PDDL plan validation, to verify plan correctness. Given a domain specification, problem instance, and proposed action sequence, VAL checks:
\begin{itemize}
    \item Each action's preconditions are satisfied when it is executed.
    \item The final state after executing all actions satisfies the goal.
\end{itemize}

VAL provides a binary validity judgment and, for invalid plans, identifies the first action whose preconditions fail.

\subsubsection{Optimality Verification}

To assess plan quality, we compute optimal solutions using the Fast Downward planning system~\citep{helmert2006fast}. Fast Downward is a state-of-the-art classical planner that guarantees optimal solutions when configured with admissible heuristics. We use the A* search algorithm with the LM-cut heuristic.

For each problem, we:
\begin{enumerate}
    \item Run Fast Downward to obtain the optimal plan length.
    \item Compare the LLM's plan length against this baseline.
    \item Mark plans as optimal if lengths match and the plan is valid.
\end{enumerate}

\subsubsection{Step-by-Step Simulation}

Beyond end-to-end validation, we simulate plan execution step-by-step using proxy implementations of each subroutine. This enables:

\begin{enumerate}
    \item \textbf{First-failure identification}: We identify the exact step where the LLM's trace first diverges from ground truth, enabling localized correction generation.
    
    \item \textbf{Fine-grained error categorization}: We distinguish between:
    \begin{itemize}
        \item \emph{Applicability errors}: Proposing an action not in the applicable set.
        \item \emph{Optimality errors}: Proposing an applicable but suboptimal action.
    \end{itemize}
    
    \item \textbf{Correction generation}: For each error type, we generate the corresponding correction by querying the oracle for the correct output.
\end{enumerate}

Algorithm~\ref{alg:eval_plan} provides pseudocode for the step-by-step evaluation procedure.

  \begin{algorithm}[tb]
  \caption{Localization of Corrections on a Generated Plan}
  \label{alg:eval_plan}
  \small
  \begin{algorithmic}
  \REQUIRE Domain $\mathcal{D}$, problem $P$, predicted actions
  $[a_1,\ldots,a_n]$, oracle $\mathcal{O}$,
auxiliary-guidance flag $q_{\mathcal{D}}$
  \ENSURE Evaluation result with localized corrections
  \STATE $s\gets\mathcal{O}.\text{extract\_initial\_state}(P)$
  \STATE $g\gets\mathcal{O}.\text{extract\_goal}(P)$
  \STATE corrections $\gets []$
  \STATE first\_invalid $\gets$ null
  \STATE first\_guidance\_error $\gets$ null
  \FOR{$i=1$ to $n$}
      \STATE $A_{\mathrm{app}}\gets
          \mathcal{O}.\text{get\_applicable\_actions}(s,g)$
      \IF{$a_i\notin A_{\mathrm{app}}$}
          \STATE first\_invalid $\gets i$
          \STATE corrections.append$(
              \text{``get\_applicable\_actions''},
              (s,g),A_{\mathrm{app}})$
          \STATE \textbf{break}
          \hfill $\triangleright$ Later states are undefined
      \ENDIF
      \IF{$q_{\mathcal{D}}$ \AND
          first\_guidance\_error is null}
          \STATE $A_{\mathrm{guide}}\gets
              \mathcal{O}.\text{get\_optimal\_actions}(s,g)$
          \IF{$a_i\notin A_{\mathrm{guide}}$}
              \STATE first\_guidance\_error $\gets i$
              \STATE corrections.append$(
                  \text{``get\_optimal\_actions''},
                  (s,g),A_{\mathrm{guide}})$
          \ENDIF
      \ENDIF
      \STATE $s\gets\mathcal{O}.\text{apply\_action}(s,a_i)$
  \ENDFOR
  \STATE goal\_reached $\gets\mathcal{O}.\text{at\_goal}(s,g)$
  \STATE \textbf{return}
  $\{$first\_invalid, first\_guidance\_error,
  corrections, goal\_reached$\}$
  \end{algorithmic}
  \end{algorithm}
We set $q_{\mathcal{D}}=\mathrm{true}$ for Gridworld, Maze, Sokoban Grid, and Full Sokoban, and $q_{\mathcal{D}}=\mathrm{false}$ for BlocksWorld. The main paper analyzes applicability errors; the auxiliary feedback is included here to specify the training procedure completely.

\subsection{Oracle Implementation}
\label{app:oracle}

The oracle provides ground-truth outputs for each subroutine. We implement oracles using a combination of external planning tools and logic-based simulation.

\paragraph{Planning Tools.} We use Fast Downward~\citep{helmert2006fast} for optimal plan computation and action applicability. For domains requiring multiple optimal plans (to compute optimal action sets), we additionally use the K* planner~\citep{katz-lee-ijcai2023}, which enumerates the top-$k$ shortest plans.

\paragraph{State Simulation.} Action effects are computed using the Tarski planning library~\citep{tarski:github:18}, which provides PDDL parsing and grounded action simulation. Given a PDDL domain and problem, Tarski computes:
\begin{itemize}
    \item The set of ground actions applicable in any state.
    \item The successor state resulting from applying an action.
    \item Whether a state satisfies a goal formula.
\end{itemize}

\paragraph{Optimality Computation.} Computing optimal actions (those on \emph{some} optimal path) requires enumerating multiple optimal plans. We use K* to generate all plans of optimal length, then take the union of first actions across these plans. For efficiency, we cache optimal action sets for frequently-queried states.

\subsection{Prompt Construction}
\label{app:prompt_construction}

The final prompt sent to the LLM consists of four components assembled in sequence:

\begin{enumerate}
    \item \textbf{Task description}: Natural language explanation of the planning domain, valid actions, and objective.
    
    \item \textbf{Subroutine documentation}: For each subroutine, we include:
    \begin{itemize}
        \item Function signature with typed arguments and return type.
        \item Natural language description of the subroutine's purpose.
        \item Accumulated L-ICL corrections as doctest examples.
    \end{itemize}
    
    \item \textbf{Example traces}: A small number ($k=2$--$3$) of complete reasoning traces showing how subroutines are invoked to solve example problems.
    
    \item \textbf{Query problem}: The problem instance to solve, formatted consistently with the examples, followed by instructions to produce a trace.
\end{enumerate}

\paragraph{State Representation.} For grid-based domains, we evaluate two state representations:
\begin{itemize}
    \item \textbf{Textual}: Positions as coordinates (e.g., ``agent at (3,4)'') with walls listed explicitly.
    \item \textbf{ASCII}: Visual grid representation where walls are marked characters and open cells are spaces.
\end{itemize}

Our ablation (Section~\ref{sec:results-grid-ablation}) shows that L-ICL achieves comparable peak performance with either representation, though ASCII grids accelerate early learning.

\subsection{Experimental Infrastructure}
\label{app:infrastructure}

\paragraph{Hardware.} Experiments were conducted on a Windows 11 Pro workstation with 64GB RAM, on a WSL2 (Ubuntu) environment. External planning tools (Fast Downward, VAL) were run locally. LLM inference was performed via API calls. The machine used an Intel Core Ultra 7 165U processor. The experiments can be carried out on ordinary consumer hardware.

\paragraph{LLM Services.} We evaluated models through their respective APIs:
\begin{itemize}
    \item DeepSeek V3 and V3.1 via the DeepSeek API.
    \item Claude Haiku 4.5 and Claude Sonnet 4.5 via the Anthropic API.
\end{itemize}

\paragraph{Hyperparameters.} Unless otherwise specified:
\begin{itemize}
    \item Temperature: 1.0.
    \item Maximum generation length: 32000 tokens.
    \item Training examples per iteration: 10 (10 problems per L-ICL update).
    \item Total training problems: up to 240.
    \item Thinking tokens for Sonnet 4.5: 10k
    \item Thinking tokens for Haiku 4.5: 5k
The hyperparameters were set according to expected maximum length of paths, balancing cost and time. 8000, 16000, 64000 tokens were also tried; we observed the longest trajectories tended to be <25,000 tokens and set parameters accordingly.
\end{itemize}

\paragraph{Timeout Handling.} Fast Downward was given a 60-second timeout per problem. Problems exceeding this limit were marked as having unknown optimal cost and excluded from optimality statistics (but included in validity statistics if the validator succeeded). DeepSeek-V3 displayed a peculiar behaviour, which sometimes occurred in DeepSeek v3.1, but no other model family - every time the character \extremeglyph{} would appear, all subsequent outputs would merely be repetitions of this token until max token limit. No other model displayed this behaviour, as such every time this character appeared in the output, the response was discarded and retried uptil 50 retries.

\paragraph{Libraries.}
The entire codebase is in Python, with the precise libraries being provided in the codebase release with this paper.


\section{Representative Prompts}
\label{app:prompts}

This appendix provides representative prompts used in all experiments. We organize prompts into two categories: (1)  L-ICL prompts based on Program Trace Prompting (PTP), and (2) baseline method prompts used for comparison approaches. All prompts use template variables (denoted with curly braces, e.g., \texttt{\{partial\_program\}}) that are replaced with problem-specific content at runtime.

\subsection{L-ICL Prompts (Program Trace Prompting)}
\label{app:licl_prompts}

L-ICL prompts follow the Program Trace Prompting (PTP) format, where the LLM is asked to predict the output of a partially specified program. The key insight is that by withholding subroutine implementations (replacing them with ``...'' markers), the LLM must infer correct behavior from documentation and accumulated examples.

\subsubsection{Base L-ICL Prompt (No Domain Visualization)}
\label{app:licl_base}

This is the minimal L-ICL prompt used for gridworld and Sokoban navigation tasks when no ASCII grid visualization is provided. The LLM must infer spatial constraints purely from accumulated L-ICL corrections.

\begin{lstlisting}[basicstyle=\footnotesize\ttfamily,breaklines=true]
Consider the program fragment below. This program fragment is 
incomplete, with key parts of the implementation hidden, by 
replacing them with "..." markers.

PROGRAM:
```python
{partial_program}
```

QUESTION: Predict what the output of the program above will be, 
given the input shown below.

Respond with the FULL program output, and ONLY the expected 
program output: you will be PENALIZED if you introduce any 
additional explanatory text.

```
>>> {task_name}({input_str})
```
\end{lstlisting}

\paragraph{Template Variables.}
\begin{itemize}
    \item \texttt{\{partial\_program\}}: The PTP-style program with subroutine signatures, documentation, doctest examples (including L-ICL corrections), and ``...'' placeholders for implementations.
    \item \texttt{\{task\_name\}}: The function name to invoke (e.g., \texttt{solve\_gridworld}).
    \item \texttt{\{input\_str\}}: The problem specification as a string (e.g., start position, goal position, wall locations).
\end{itemize}

\subsubsection{L-ICL Prompt with ASCII Grid Visualization}
\label{app:licl_grid}

When ASCII grid visualization is enabled, the prompt includes a visual representation of the environment. This provides spatial scaffolding that accelerates early learning, though L-ICL achieves comparable peak performance without it.

\begin{lstlisting}[basicstyle=\footnotesize\ttfamily,breaklines=true]
Consider the program fragment below. This program fragment is 
incomplete, with key parts of the implementation hidden, by 
replacing them with "..." markers.

IMPORTANT: You are an agent navigating a {grid_size} gridworld.
The grid has {num_walls} walls that block movement.

**Grid Layout:**
```
     1   2   3   4   5   6   7   8   9   10
   +---+---+---+---+---+---+---+---+---+---+
10 | . | . | . | . | . | . | . | . | . | . |
   +---+---+---+---+---+---+---+---+---+---+
 9 | . | . | # | # | # | # | # | # | # | . |
   +---+---+---+---+---+---+---+---+---+---+
 8 | . | # | . | # | . | # | . | . | . | . |
   +---+---+---+---+---+---+---+---+---+---+
 7 | . | # | . | . | . | # | . | # | # | . |
   +---+---+---+---+---+---+---+---+---+---+
 6 | . | # | . | # | . | . | . | # | . | . |
   +---+---+---+---+---+---+---+---+---+---+
 5 | . | . | . | # | . | # | . | # | . | . |
   +---+---+---+---+---+---+---+---+---+---+
 4 | . | # | . | # | . | # | . | . | # | . |
   +---+---+---+---+---+---+---+---+---+---+
 3 | . | # | . | . | . | # | . | # | . | . |
   +---+---+---+---+---+---+---+---+---+---+
 2 | . | # | . | # | . | . | . | # | . | . |
   +---+---+---+---+---+---+---+---+---+---+
 1 | . | . | . | . | . | . | . | . | . | . |
   +---+---+---+---+---+---+---+---+---+---+
```

PROGRAM:
```python
{partial_program}
```

QUESTION: Predict what the output of the program above will be, 
given the input shown below.

Respond with the FULL program output, and ONLY the expected 
program output: you will be PENALIZED if you introduce any 
additional explanatory text.

```
>>> {task_name}({input_str})
```
\end{lstlisting}

\paragraph{Grid Symbols.}
\begin{itemize}
    \item \texttt{.} -- Open cell (traversable)
    \item \texttt{\#} -- Wall (impassable)
    \item \texttt{\$} -- Box (in Sokoban)
\end{itemize}

\subsubsection{L-ICL BlocksWorld Prompt with UBW Algorithm}
\label{app:licl_blocksworld}

For BlocksWorld, we additionally provide algorithmic guidance based on the Universal Blocks World (UBW) algorithm~\citep{stechly2024chain}. This tests whether L-ICL can improve adherence to prescribed planning strategies beyond simple constraint satisfaction.

\begin{lstlisting}[basicstyle=\footnotesize\ttfamily,breaklines=true]
Consider the program fragment below. This program fragment 
implements the Universal Blocks World (UBW) algorithm, which is 
a systematic two-phase approach for solving blocks world planning 
problems. The implementation is incomplete, with key parts 
replaced by "..." markers.

UNIVERSAL BLOCKS WORLD ALGORITHM OVERVIEW:
The UBW algorithm works in two distinct phases to efficiently 
solve any blocks world configuration:

PHASE 1: STRATEGIC UNSTACKING
- Unstack ALL blocks that are stacked on top of others
- Work from top to bottom, unstacking clear blocks first
- Move incorrectly positioned blocks to the table

PHASE 2: SYSTEMATIC REASSEMBLY
- Build goal configurations from bottom up
- Process blocks in dependency order (place supporting blocks 
  before supported blocks)
- Only place a block when its target is ready (clear and in 
  final position)
- Ensure structural integrity throughout construction

KEY HEURISTICS FOR IMPLEMENTATION:

1. STATE ANALYSIS:
   - Parse predicates into on(), on-table(), and clear() 
     relationships
   - Build dependency graphs: what should be on what
   - Identify bottom blocks (blocks that should be on table 
     in goal)

2. UNSTACKING STRATEGY:
   - Check each on(X,Y) relationship in current state
   - If (X,Y) is NOT in goal relationships, consider 
     unstacking X
   - Only unstack if X is clear (no blocks on top)
   - Priority: unstack blocks that block other necessary moves

3. REASSEMBLY STRATEGY:
   - For each goal on(X,Y), check if X can be placed on Y
   - X must be: clear AND on-table
   - Y must be: clear AND in its final position
   - Y is in final position if: Y should be on table OR Y is 
     already correctly placed on its target

4. ACTION SELECTION LOGIC:
   ```
   For unstacking: if on(X,Y) in current state AND clear(X):
       return move-b-to-t(X,Y)
   
   For assembly: if goal requires on(X,Y) AND 
                    can_place_block(X,Y):
       return move-t-to-b(X,Y)
   ```

5. CORRECTNESS VERIFICATION:
   - Always verify preconditions before suggesting actions
   - Check that actions don't break existing correct 
     configurations
   - Ensure goal-directed progress in every move during 
     assembly phase

DETAILED TRACE GUIDANCE:
When implementing the UBW algorithm, provide step-by-step 
reasoning inside reasoning() calls if required, which is your 
scratchpad.

1. State the current configuration clearly
2. Identify which phase you're in (unstacking vs assembly)
3. Explain WHY each action is chosen based on UBW principles
4. Show how the action advances toward the goal
5. Verify preconditions are satisfied
6. Update state representation after each action

PROGRAM:
```python
{partial_program}
```

QUESTION: Predict what the output of the program above will be, 
given the input shown below.

IMPLEMENTATION REQUIREMENTS:
- Follow the UBW algorithm phases strictly
- Provide detailed reasoning for each action selection
- Show state analysis and dependency tracking
- Explain how each move contributes to the overall strategy
- Demonstrate understanding of when to unstack vs when to build
- Verify that all actions follow UBW heuristics

Respond with the FULL program output, including detailed 
algorithmic traces that demonstrate proper UBW implementation. 
Your trace should show:
- Clear identification of current phase (unstacking/assembly)
- Specific reasoning for each action choice
- State updates and goal progress tracking
- Verification that actions follow UBW principles

Under no circumstance must you skip steps in the program output. 
You CAN decide to go back and choose different actions if you 
feel that you have made a mistake, but the FINAL PLAN must show 
the COMPLETE CORRECT PATH ONLY.

```
>>> {task_name}({input_str})
```
\end{lstlisting}

\subsubsection{Example Partial Program Structure}
\label{app:partial_program}

The \texttt{\{partial\_program\}} template variable is replaced with a PTP-style program containing subroutine documentation and accumulated L-ICL corrections. Below is a representative example for gridworld navigation:

\begin{lstlisting}[basicstyle=\footnotesize\ttfamily,breaklines=true]
import collections
from typing import Dict, List, Set, Tuple, Union, Optional, Any, FrozenSet

PlanningState = Any
Action = Any


@traced
def extract_problem(input_str: str) -> str:
    """Extract a standardized problem description from input.

    """
    ...

@traced
def extract_initial_state(problem_str: str) -> PlanningState:
    """Extract the initial state from a problem description.

    """
    ...

@traced
def extract_goal(problem_str: str) -> PlanningState:
    """Extract the goal from a problem description.

    """
    ...

@traced
def at_goal(state: PlanningState, goal: PlanningState) -> bool:
    """Check if current state satisfies goal conditions.

    """
    ...

@traced
def get_applicable_actions(state: PlanningState, goal: PlanningState) -> Set[Action]:
    """Get all applicable actions in the current state.

    """
    ...

@traced
def get_optimal_actions(state: PlanningState, applicable_actions: List[Action],
                      goal: PlanningState) -> Set[Action]:
    """Get actions that are part of an optimal plan.

    """
    ...

@traced
def apply_action(state: PlanningState, action: Action, goal: PlanningState) -> PlanningState:
    """Apply an action to a state, returning the resulting state.

    """
    ...

def pddl_grid(input_str: str):
    """Solve a planning problem described in input_str.

    This function processes a planning problem description by:
    1. Extracting the initial state and goal
    2. Iteratively applying actions until the goal is reached
    3. Returning the sequence of actions as a plan

    >>> pddl_grid('(define (problem gw-task-351)\n  (:domain gridworld-10x10)\n  (:init (at c9-5))\n  (:goal (at c5-10))\n)\n')
    Calling extract_problem('(define (problem gw-task-351)\n  (:domain gridworld-10x10)\n  (:init (at c9-5))\n  (:goal (at c5-10))\n)\n')...
    ...extract_problem returned 'gridworld-10x10'
    Calling extract_initial_state('(define (problem gw-task-351)\n  (:domain gridworld-10x10)\n  (:init (at c9-5))\n  (:goal (at c5-10))\n)\n')...
    ...extract_initial_state returned (9, 5)
    Calling extract_goal('(define (problem gw-task-351)\n  (:domain gridworld-10x10)\n  (:init (at c9-5))\n  (:goal (at c5-10))\n)\n')...
    ...extract_goal returned (5, 10)
    Calling at_goal((9, 5), (5, 10))...
    ...at_goal returned False
    Calling get_applicable_actions((9, 5), (5, 10))...
    ...get_applicable_actions returned ['move-north', 'move-east']
    Calling get_optimal_actions((9, 5), ['move-north', 'move-east'], (5, 10))...
    ...get_optimal_actions returned ['move-north', 'move-east']
    Calling apply_action((9, 5), 'move-north', (5, 10))...
    ...apply_action returned (9, 6)
    Calling at_goal((9, 6), (5, 10))...
    ...at_goal returned False
    Calling get_applicable_actions((9, 6), (5, 10))...
    ...get_applicable_actions returned ['move-south', 'move-east']
    Calling get_optimal_actions((9, 6), ['move-south', 'move-east'], (5, 10))...
    ...get_optimal_actions returned ['move-east']
    Calling apply_action((9, 6), 'move-east', (5, 10))...
    ...apply_action returned (10, 6)
    Calling at_goal((10, 6), (5, 10))...
    ...at_goal returned False
    Calling get_applicable_actions((10, 6), (5, 10))...
    ...get_applicable_actions returned ['move-north', 'move-south', 'move-west']
    Calling get_optimal_actions((10, 6), ['move-north', 'move-south', 'move-west'], (5, 10))...
    ...get_optimal_actions returned ['move-north']
    Calling apply_action((10, 6), 'move-north', (5, 10))...
    ...apply_action returned (10, 7)
    Calling at_goal((10, 7), (5, 10))...
    ...at_goal returned False
    Calling get_applicable_actions((10, 7), (5, 10))...
    ...get_applicable_actions returned ['move-north', 'move-south']
    Calling get_optimal_actions((10, 7), ['move-north', 'move-south'], (5, 10))...
    ...get_optimal_actions returned ['move-north']
    Calling apply_action((10, 7), 'move-north', (5, 10))...
    ...apply_action returned (10, 8)
    Calling at_goal((10, 8), (5, 10))...
    ...at_goal returned False
    Calling get_applicable_actions((10, 8), (5, 10))...
    ...get_applicable_actions returned ['move-north', 'move-south', 'move-west']
    Calling get_optimal_actions((10, 8), ['move-north', 'move-south', 'move-west'], (5, 10))...
    ...get_optimal_actions returned ['move-north']
    Calling apply_action((10, 8), 'move-north', (5, 10))...
    ...apply_action returned (10, 9)
    Calling at_goal((10, 9), (5, 10))...
    ...at_goal returned False
    Calling get_applicable_actions((10, 9), (5, 10))...
    ...get_applicable_actions returned ['move-north', 'move-south']
    Calling get_optimal_actions((10, 9), ['move-north', 'move-south'], (5, 10))...
    ...get_optimal_actions returned ['move-north']
    Calling apply_action((10, 9), 'move-north', (5, 10))...
    ...apply_action returned (10, 10)
    Calling at_goal((10, 10), (5, 10))...
    ...at_goal returned False
    Calling get_applicable_actions((10, 10), (5, 10))...
    ...get_applicable_actions returned ['move-south', 'move-west']
    Calling get_optimal_actions((10, 10), ['move-south', 'move-west'], (5, 10))...
    ...get_optimal_actions returned ['move-west']
    Calling apply_action((10, 10), 'move-west', (5, 10))...
    ...apply_action returned (9, 10)
    Calling at_goal((9, 10), (5, 10))...
    ...at_goal returned False
    Calling get_applicable_actions((9, 10), (5, 10))...
    ...get_applicable_actions returned ['move-east', 'move-west']
    Calling get_optimal_actions((9, 10), ['move-east', 'move-west'], (5, 10))...
    ...get_optimal_actions returned ['move-west']
    Calling apply_action((9, 10), 'move-west', (5, 10))...
    ...apply_action returned (8, 10)
    Calling at_goal((8, 10), (5, 10))...
    ...at_goal returned False
    Calling get_applicable_actions((8, 10), (5, 10))...
    ...get_applicable_actions returned ['move-east', 'move-west']
    Calling get_optimal_actions((8, 10), ['move-east', 'move-west'], (5, 10))...
    ...get_optimal_actions returned ['move-west']
    Calling apply_action((8, 10), 'move-west', (5, 10))...
    ...apply_action returned (7, 10)
    Calling at_goal((7, 10), (5, 10))...
    ...at_goal returned False
    Calling get_applicable_actions((7, 10), (5, 10))...
    ...get_applicable_actions returned ['move-east', 'move-west']
    Calling get_optimal_actions((7, 10), ['move-east', 'move-west'], (5, 10))...
    ...get_optimal_actions returned ['move-west']
    Calling apply_action((7, 10), 'move-west', (5, 10))...
    ...apply_action returned (6, 10)
    Calling at_goal((6, 10), (5, 10))...
    ...at_goal returned False
    Calling get_applicable_actions((6, 10), (5, 10))...
    ...get_applicable_actions returned ['move-east', 'move-west']
    Calling get_optimal_actions((6, 10), ['move-east', 'move-west'], (5, 10))...
    ...get_optimal_actions returned ['move-west']
    Calling apply_action((6, 10), 'move-west', (5, 10))...
    ...apply_action returned (5, 10)
    Calling at_goal((5, 10), (5, 10))...
    ...at_goal returned True
    Final answer: move-north move-east move-north move-north move-north move-north move-west move-west move-west move-west move-west
    ['move-north', 'move-east', 'move-north', 'move-north', 'move-north', 'move-north', 'move-west', 'move-west', 'move-west', 'move-west', 'move-west']

    >>> pddl_grid('(define (problem gw-task-352)\n  (:domain gridworld-10x10)\n  (:init (at c9-3))\n  (:goal (at c7-7))\n)\n')
    Calling extract_problem('(define (problem gw-task-352)\n  (:domain gridworld-10x10)\n  (:init (at c9-3))\n  (:goal (at c7-7))\n)\n')...
    ...extract_problem returned 'gridworld-10x10'
    Calling extract_initial_state('(define (problem gw-task-352)\n  (:domain gridworld-10x10)\n  (:init (at c9-3))\n  (:goal (at c7-7))\n)\n')...
    ...extract_initial_state returned (9, 3)
    Calling extract_goal('(define (problem gw-task-352)\n  (:domain gridworld-10x10)\n  (:init (at c9-3))\n  (:goal (at c7-7))\n)\n')...
    ...extract_goal returned (7, 7)
    Calling at_goal((9, 3), (7, 7))...
    ...at_goal returned False
    Calling get_applicable_actions((9, 3), (7, 7))...
    ...get_applicable_actions returned ['move-south', 'move-east']
    Calling get_optimal_actions((9, 3), ['move-south', 'move-east'], (7, 7))...
    ...get_optimal_actions returned ['move-south', 'move-east']
    Calling apply_action((9, 3), 'move-south', (7, 7))...
    ...apply_action returned (9, 2)
    Calling at_goal((9, 2), (7, 7))...
    ...at_goal returned False
    Calling get_applicable_actions((9, 2), (7, 7))...
    ...get_applicable_actions returned ['move-north', 'move-south', 'move-east']
    Calling get_optimal_actions((9, 2), ['move-north', 'move-south', 'move-east'], (7, 7))...
    ...get_optimal_actions returned ['move-south']
    Calling apply_action((9, 2), 'move-south', (7, 7))...
    ...apply_action returned (9, 1)
    Calling at_goal((9, 1), (7, 7))...
    ...at_goal returned False
    Calling get_applicable_actions((9, 1), (7, 7))...
    ...get_applicable_actions returned ['move-north', 'move-east', 'move-west']
    Calling get_optimal_actions((9, 1), ['move-north', 'move-east', 'move-west'], (7, 7))...
    ...get_optimal_actions returned ['move-west']
    Calling apply_action((9, 1), 'move-west', (7, 7))...
    ...apply_action returned (8, 1)
    Calling at_goal((8, 1), (7, 7))...
    ...at_goal returned False
    Calling get_applicable_actions((8, 1), (7, 7))...
    ...get_applicable_actions returned ['move-east', 'move-west']
    Calling get_optimal_actions((8, 1), ['move-east', 'move-west'], (7, 7))...
    ...get_optimal_actions returned ['move-west']
    Calling apply_action((8, 1), 'move-west', (7, 7))...
    ...apply_action returned (7, 1)
    Calling at_goal((7, 1), (7, 7))...
    ...at_goal returned False
    Calling get_applicable_actions((7, 1), (7, 7))...
    ...get_applicable_actions returned ['move-north', 'move-east', 'move-west']
    Calling get_optimal_actions((7, 1), ['move-north', 'move-east', 'move-west'], (7, 7))...
    ...get_optimal_actions returned ['move-north']
    Calling apply_action((7, 1), 'move-north', (7, 7))...
    ...apply_action returned (7, 2)
    Calling at_goal((7, 2), (7, 7))...
    ...at_goal returned False
    Calling get_applicable_actions((7, 2), (7, 7))...
    ...get_applicable_actions returned ['move-north', 'move-south', 'move-west']
    Calling get_optimal_actions((7, 2), ['move-north', 'move-south', 'move-west'], (7, 7))...
    ...get_optimal_actions returned ['move-north']
    Calling apply_action((7, 2), 'move-north', (7, 7))...
    ...apply_action returned (7, 3)
    Calling at_goal((7, 3), (7, 7))...
    ...at_goal returned False
    Calling get_applicable_actions((7, 3), (7, 7))...
    ...get_applicable_actions returned ['move-north', 'move-south']
    Calling get_optimal_actions((7, 3), ['move-north', 'move-south'], (7, 7))...
    ...get_optimal_actions returned ['move-north']
    Calling apply_action((7, 3), 'move-north', (7, 7))...
    ...apply_action returned (7, 4)
    Calling at_goal((7, 4), (7, 7))...
    ...at_goal returned False
    Calling get_applicable_actions((7, 4), (7, 7))...
    ...get_applicable_actions returned ['move-north', 'move-south', 'move-east']
    Calling get_optimal_actions((7, 4), ['move-north', 'move-south', 'move-east'], (7, 7))...
    ...get_optimal_actions returned ['move-north']
    Calling apply_action((7, 4), 'move-north', (7, 7))...
    ...apply_action returned (7, 5)
    Calling at_goal((7, 5), (7, 7))...
    ...at_goal returned False
    Calling get_applicable_actions((7, 5), (7, 7))...
    ...get_applicable_actions returned ['move-north', 'move-south']
    Calling get_optimal_actions((7, 5), ['move-north', 'move-south'], (7, 7))...
    ...get_optimal_actions returned ['move-north']
    Calling apply_action((7, 5), 'move-north', (7, 7))...
    ...apply_action returned (7, 6)
    Calling at_goal((7, 6), (7, 7))...
    ...at_goal returned False
    Calling get_applicable_actions((7, 6), (7, 7))...
    ...get_applicable_actions returned ['move-north', 'move-south', 'move-west']
    Calling get_optimal_actions((7, 6), ['move-north', 'move-south', 'move-west'], (7, 7))...
    ...get_optimal_actions returned ['move-north']
    Calling apply_action((7, 6), 'move-north', (7, 7))...
    ...apply_action returned (7, 7)
    Calling at_goal((7, 7), (7, 7))...
    ...at_goal returned True
    Final answer: move-south move-south move-west move-west move-north move-north move-north move-north move-north move-north
    ['move-south', 'move-south', 'move-west', 'move-west', 'move-north', 'move-north', 'move-north', 'move-north', 'move-north', 'move-north']

    >>> pddl_grid('(define (problem gw-task-353)\n  (:domain gridworld-10x10)\n  (:init (at c7-2))\n  (:goal (at c2-5))\n)\n')
    Calling extract_problem('(define (problem gw-task-353)\n  (:domain gridworld-10x10)\n  (:init (at c7-2))\n  (:goal (at c2-5))\n)\n')...
    ...extract_problem returned 'gridworld-10x10'
    Calling extract_initial_state('(define (problem gw-task-353)\n  (:domain gridworld-10x10)\n  (:init (at c7-2))\n  (:goal (at c2-5))\n)\n')...
    ...extract_initial_state returned (7, 2)
    Calling extract_goal('(define (problem gw-task-353)\n  (:domain gridworld-10x10)\n  (:init (at c7-2))\n  (:goal (at c2-5))\n)\n')...
    ...extract_goal returned (2, 5)
    Calling at_goal((7, 2), (2, 5))...
    ...at_goal returned False
    Calling get_applicable_actions((7, 2), (2, 5))...
    ...get_applicable_actions returned ['move-north', 'move-south', 'move-west']
    Calling get_optimal_actions((7, 2), ['move-north', 'move-south', 'move-west'], (2, 5))...
    ...get_optimal_actions returned ['move-west']
    Calling apply_action((7, 2), 'move-west', (2, 5))...
    ...apply_action returned (6, 2)
    Calling at_goal((6, 2), (2, 5))...
    ...at_goal returned False
    Calling get_applicable_actions((6, 2), (2, 5))...
    ...get_applicable_actions returned ['move-south', 'move-east', 'move-west']
    Calling get_optimal_actions((6, 2), ['move-south', 'move-east', 'move-west'], (2, 5))...
    ...get_optimal_actions returned ['move-west']
    Calling apply_action((6, 2), 'move-west', (2, 5))...
    ...apply_action returned (5, 2)
    Calling at_goal((5, 2), (2, 5))...
    ...at_goal returned False
    Calling get_applicable_actions((5, 2), (2, 5))...
    ...get_applicable_actions returned ['move-north', 'move-south', 'move-east']
    Calling get_optimal_actions((5, 2), ['move-north', 'move-south', 'move-east'], (2, 5))...
    ...get_optimal_actions returned ['move-north']
    Calling apply_action((5, 2), 'move-north', (2, 5))...
    ...apply_action returned (5, 3)
    Calling at_goal((5, 3), (2, 5))...
    ...at_goal returned False
    Calling get_applicable_actions((5, 3), (2, 5))...
    ...get_applicable_actions returned ['move-north', 'move-south', 'move-west']
    Calling get_optimal_actions((5, 3), ['move-north', 'move-south', 'move-west'], (2, 5))...
    ...get_optimal_actions returned ['move-west']
    Calling apply_action((5, 3), 'move-west', (2, 5))...
    ...apply_action returned (4, 3)
    Calling at_goal((4, 3), (2, 5))...
    ...at_goal returned False
    Calling get_applicable_actions((4, 3), (2, 5))...
    ...get_applicable_actions returned ['move-east', 'move-west']
    Calling get_optimal_actions((4, 3), ['move-east', 'move-west'], (2, 5))...
    ...get_optimal_actions returned ['move-west']
    Calling apply_action((4, 3), 'move-west', (2, 5))...
    ...apply_action returned (3, 3)
    Calling at_goal((3, 3), (2, 5))...
    ...at_goal returned False
    Calling get_applicable_actions((3, 3), (2, 5))...
    ...get_applicable_actions returned ['move-north', 'move-south', 'move-east']
    Calling get_optimal_actions((3, 3), ['move-north', 'move-south', 'move-east'], (2, 5))...
    ...get_optimal_actions returned ['move-north']
    Calling apply_action((3, 3), 'move-north', (2, 5))...
    ...apply_action returned (3, 4)
    Calling at_goal((3, 4), (2, 5))...
    ...at_goal returned False
    Calling get_applicable_actions((3, 4), (2, 5))...
    ...get_applicable_actions returned ['move-north', 'move-south']
    Calling get_optimal_actions((3, 4), ['move-north', 'move-south'], (2, 5))...
    ...get_optimal_actions returned ['move-north']
    Calling apply_action((3, 4), 'move-north', (2, 5))...
    ...apply_action returned (3, 5)
    Calling at_goal((3, 5), (2, 5))...
    ...at_goal returned False
    Calling get_applicable_actions((3, 5), (2, 5))...
    ...get_applicable_actions returned ['move-north', 'move-south', 'move-west']
    Calling get_optimal_actions((3, 5), ['move-north', 'move-south', 'move-west'], (2, 5))...
    ...get_optimal_actions returned ['move-west']
    Calling apply_action((3, 5), 'move-west', (2, 5))...
    ...apply_action returned (2, 5)
    Calling at_goal((2, 5), (2, 5))...
    ...at_goal returned True
    Final answer: move-west move-west move-north move-west move-west move-north move-north move-west
    ['move-west', 'move-west', 'move-north', 'move-west', 'move-west', 'move-north', 'move-north', 'move-west']

    """
    ...

\end{lstlisting}

\subsection{Baseline Method Prompts}
\label{app:baseline_prompts}

The following prompts are used for baseline comparison methods. All baselines receive the same problem information but use different reasoning frameworks.

\subsubsection{Zero-Shot CoT / RAG-CoT Prompt}
\label{app:cot_prompt}

The base prompt used for Zero-Shot Chain-of-Thought and RAG-CoT baselines. For RAG-CoT, dynamically retrieved examples are inserted in the \texttt{\{examples\}} section.

\begin{lstlisting}[basicstyle=\footnotesize\ttfamily,breaklines=true]
You are an expert at navigating gridworld environments. You will 
solve navigation problems where an agent must find the optimal 
path from a start position to a goal position while avoiding 
walls and obstacles.

IMPORTANT:
You are an agent navigating a {grid_size} gridworld.
The grid has {num_walls} walls that block movement.

**Grid Layout:**
{ascii_grid}

# Task Description

In each problem, you are given:
- A gridworld of specific dimensions
- A start position (row, column)
- A goal position (row, column)
- Wall locations that block movement

Your task is to find the shortest path from start to goal using 
these actions:
- **move-north**: Move one cell north (increase row by 1)
- **move-south**: Move one cell south (decrease row by 1)
- **move-east**: Move one cell east (increase column by 1)
- **move-west**: Move one cell west (decrease column by 1)

# Solution Strategy

For each problem, follow this reasoning process:

1. **Analyze the Grid**: Identify the start position, goal 
   position, and obstacles
2. **Plan the Route**: Determine if a direct path exists or if 
   you need to navigate around obstacles
3. **Step-by-Step Reasoning**: For each move, explain why it 
   brings you closer to the goal
4. **Verify the Path**: Ensure the path is valid and optimal

# Example Problems

{examples}

# Problem to Solve

Start: {start_position}
Goal: {goal_position}

{deadzone_warning}

Please solve this problem step-by-step and provide your answer.

**Your Solution:**

First, provide your step-by-step reasoning:
1. Identify the start position, goal position, and any obstacles
2. Reason through each step of your path
3. Verify your path is valid and optimal

Then, provide your final answer EXACTLY in this format:

**Final Action Sequence:** move-direction1, move-direction2, ...

IMPORTANT: You MUST include the line starting with 
"Final Action Sequence:" followed by your comma-separated list 
of actions.
\end{lstlisting}

\subsubsection{Self-Consistency Prompt}
\label{app:sc_prompt}

Self-Consistency uses the same base prompt as CoT, with a sample annotation appended to each independent call:

\begin{lstlisting}[basicstyle=\footnotesize\ttfamily,breaklines=true]
{base_cot_prompt}

<!-- Self-Consistency Sample {k}/{total}: Treat this run 
independently and produce a complete plan -->
\end{lstlisting}

Each sample is generated with temperature $> 0$ for diversity. The final answer is selected via majority voting over the \texttt{k} samples.

\subsubsection{Self-Refine Refinement Prompt}
\label{app:refine_prompt}

After the initial attempt, if refinement rounds remain, the model receives its previous response with reflection instructions:

\begin{lstlisting}[basicstyle=\footnotesize\ttfamily,breaklines=true]
{base_cot_prompt}

### Self-Refinement Attempt {attempt_number}
You previously produced the following reasoning and plan:

{previous_response}

Proposed action sequence: {previous_actions}

Carefully re-read the task description and your earlier steps. 
Without running code or simulations, check for potential 
mistakes:
- Did any move leave the grid or pass through a wall?
- Does the sequence actually reach the goal cell?
- Is there a shorter valid route?

If issues are found, explain them briefly and provide a 
corrected plan. If you believe the plan is correct and needs 
no further refinement, explicitly state:
'**No further refinement needed.**' and then restate the 
action sequence.

Always finish with a line of the form:
**Final Action Sequence:** move-*, move-*, ...

Refined solution:
\end{lstlisting}

\subsubsection{ReAct (Prompt-Only) Prompt}
\label{app:react_prompt}

ReAct uses an alternating Thought/Action trace format:

\begin{lstlisting}[basicstyle=\footnotesize\ttfamily,breaklines=true]
You are an expert gridworld planner. Solve using ReAct style 
trace.

IMPORTANT:
You are an agent navigating a {grid_size} gridworld.
The grid has {num_walls} walls that block movement.

**Grid Layout:**
{ascii_grid}

## Valid Actions
- **move-north**: Move one cell up (increase y by 1)
- **move-south**: Move one cell down (decrease y by 1)
- **move-east**: Move one cell right (increase x by 1)
- **move-west**: Move one cell left (decrease x by 1)

## Movement Constraints
- You cannot move through walls
- You cannot move outside the grid boundaries
- Each action moves exactly one cell

# Example

Start: (2,1), Goal: (5,4)

Thought: I am at (2,1) and need to reach (5,4). I should move 
north and east while checking for obstacles.
Action: move-north
Thought: Now at (2,2). Continue moving toward the goal.
Action: move-north
...
Final Thought: Reached the goal at (5,4).
**Final Action Sequence:** move-north, move-north, move-east, ...

# Problem to Solve

Start: {start_position}
Goal: {goal_position}

Guidelines:
- Alternate between `Thought:` and `Action:`
- Keep moves consistent with grid layout
- Avoid illegal steps (walls, boundaries)
- End with `Final Thought:` and `**Final Action Sequence:**`
\end{lstlisting}

\subsubsection{Tree-of-Thoughts Expansion Prompt}
\label{app:tot_prompt}

ToT uses a structured expansion prompt requesting JSON-formatted candidates:

\begin{lstlisting}[basicstyle=\footnotesize\ttfamily,breaklines=true]
{reference_examples}

Gridworld planning problem:
Start: {start_position}
Goal: {goal_position}

Current depth: {depth}/{max_depth}
Actions chosen so far: {action_prefix}
Thoughts considered so far:
{thought_history}

Generate up to 5 candidate expansions as JSON. Each must include:
  - "thought": a short description of the idea
  - "proposed_actions": list of up to 8 moves continuing the plan
  - "confidence": integer 0-100 for promise of success
  - "is_terminal": true if plan should stop after these actions
  - "final_plan": optional full action list if terminal

Moves must stay within bounds and avoid walls.

Return ONLY the JSON array; no commentary.
\end{lstlisting}

\subsubsection{ReAct (+Oracle) Feedback Prompt}
\label{app:oracle_prompt}

When the oracle detects errors, it provides specific feedback:

\begin{lstlisting}[basicstyle=\footnotesize\ttfamily,breaklines=true]
{original_prompt}

---

Your previous response:
{previous_response}

---

ORACLE FEEDBACK: Your plan has a PROBLEM at step {step_number}.

The action '{failed_action}' at position {position} is INVALID 
because {reason}.

Please find an alternative path that avoids this issue.

**Corrected Final Action Sequence:**
\end{lstlisting}

\paragraph{Feedback Types.}
\begin{itemize}
    \item \textbf{Invalid Move}: ``The action `move-X' at position $(x,y)$ is INVALID because it would move into a wall or out of bounds.''
    \item \textbf{Deadzone Entry}: ``The action `move-X' leads to position $(x,y)$ which is a DEADZONE. You should avoid deadzones.''
    \item \textbf{Incomplete Path}: ``Your plan is INCOMPLETE. After executing all actions, you ended at $(x,y)$ but did not reach the goal.''
\end{itemize}

\subsection{L-ICL Correction Format}
\label{app:correction_format_details}
L-ICL corrections are formatted as doctest-style input-output examples inserted into subroutine documentation. This format leverages Python's doctest convention, which is well-represented in LLM training data.

\paragraph{Correction Structure.}\mbox{}
\begin{lstlisting}[basicstyle=\footnotesize\ttfamily,breaklines=true]
>>> {function_name}({input_args})
{correct_output}
\end{lstlisting}

\paragraph{Example Corrections by Subroutine.}\mbox{}\\
\textbf{Applicability Correction} (when LLM proposes invalid action):
\begin{lstlisting}[basicstyle=\footnotesize\ttfamily,breaklines=true]
>>> get_applicable_actions(state=(3, 4), goal=(7, 8))
{'move_north', 'move_south', 'move_west'}
\end{lstlisting}

\noindent\textbf{Auxiliary Optimal-Action Correction} (when LLM proposes an applicable but suboptimal):
\begin{lstlisting}[basicstyle=\footnotesize\ttfamily,breaklines=true]
>>> get_optimal_actions(state=(5, 2), goal=(8, 7))
{'move_north', 'move_east'}
\end{lstlisting}

\subsection{Action Parsing Patterns}
\label{app:parsing}

All methods use the same action sequence parser that handles multiple output formats:

\begin{lstlisting}[language=Python,basicstyle=\footnotesize\ttfamily,breaklines=true]
PATTERNS = [
    r'\*\*Final Action Sequence:\*\*\s*(.+?)(?:\n|$)',
    r'Final Action Sequence:\s*(.+?)(?:\n|$)',
    r'\*\*Action Sequence:\*\*\s*(.+?)(?:\n|$)',
    r'Action Sequence:\s*(.+?)(?:\n|$)',
    r'Optimal path:\s*(.+?)(?:\n|$)',
    r'Plan:\s*(.+?)(?:\n|$)',
]

VALID_ACTIONS = {
    'move-north', 'move-south', 'move-east', 'move-west',
    'move_north', 'move_south', 'move_east', 'move_west',
    'push-north', 'push-south', 'push-east', 'push-west',
}

# Normalize action format
ACTION_ALIASES = {
    'north': 'move-north', 'south': 'move-south',
    'east': 'move-east', 'west': 'move-west',
    'up': 'move-north', 'down': 'move-south',
    'right': 'move-east', 'left': 'move-west',
}
\end{lstlisting}

\subsection{Prompt Variations by Domain}
\label{app:domain_variations}

\paragraph{8$\times$8 Two-Room Gridworld.} Uses the standard L-ICL prompt with ASCII grid showing two rooms separated by a wall with a doorway.

\paragraph{10$\times$10 Maze.} Uses the L-ICL prompt with ASCII grid showing procedurally generated maze corridors. Wall density is higher, creating narrow passages.

\paragraph{Sokoban-Style Gridworld.} Uses the standard L-ICL prompt with Sokoban-style ASCII layouts but no pushable boxes.

\paragraph{Full Sokoban.} Extends the action space to include push actions:
\begin{lstlisting}[basicstyle=\footnotesize\ttfamily,breaklines=true]
Valid actions: 
- Movement: move_north, move_south, move_east, move_west
- Pushing: push_north, push_south, push_east, push_west
\end{lstlisting}

\paragraph{BlocksWorld.} Uses the UBW algorithm prompt (Section~\ref{app:licl_blocksworld}) with relational state representation instead of spatial coordinates.

\subsection{Hyperparameter Settings by Method}
\label{app:hyperparams}

Table~\ref{tab:prompt_hyperparams} summarizes the key hyperparameters used for each prompting method. The hyperparameters for LLM calls were maintained as standard with the PTP LLM calls. In the cases when temperatures were not 1, we observed that at T=1 the output quality was poor and experimented in the range of T={0.0 to 0.9} in increments of 0.1.

\begin{table}[t]
\centering
{
\small
\setlength{\tabcolsep}{1mm}
\begin{tabular}{@{}lccc@{}}
\toprule
\textbf{Method} & \textbf{Temperature} & \textbf{Max Tokens} & \textbf{Samples/Iterations} \\
\midrule
Zero-Shot CoT & 1.0 & 32,000 & 1 \\
RAG-CoT & 1.0 & 32,000 & 1 \\
Self-Consistency & 1.0 & 32,000 & $k=5$ \\
Self-Refine & 1.0 & 32,000 & $N=5$ \\
ReAct (Prompt) & 1.0 & 32,000 & 1 \\
ToT (Prompt) & 1.0 & 32,000 & $b=5, d=3$ \\
ReAct (+Oracle) & 0.3 & 32,000 & 1--2 \\
\midrule
L-ICL & 1.0 & 32,000 & 1 \\
\bottomrule
\end{tabular}
}
\caption{Hyperparameter settings for each prompting method.}
\label{tab:prompt_hyperparams}
\end{table}

\paragraph{L-ICL Training Configuration.}
  \begin{itemize}
      \item Training examples: up to 240 problems.
      \item \texttt{get\_applicable\_actions} in every domain, with \texttt{get\_optimal\_actions} additionally enabled in Gridworld, Maze, Sokoban Grid, and Full Sokoban.
      \item Corrections per problem: up to two: the first auxiliary optimal-action error and the first applicability error in the navigation and Sokoban domains, and up to one, the first applicability error, in BlocksWorld.
      \item Correction accumulation: batch update after 10 training examples.
  \end{itemize}

\section{Statistical Significance and Uncertainty}
\label{app:statistical-significance}

The unit of paired inference is the held-out planning problem. When two
conditions were evaluated on the same ordered problem IDs, we used the exact
two-sided McNemar test. For three or more
repeated conditions, we first used Cochran's Q; pairwise McNemar follow-ups were carried out when appropriate, except for explicitly
predeclared two-condition controls. We report Holm-adjusted p-values within
each declared metric/setting family. Validity ($V$) is the absence of an
inapplicable action, and success ($S$) additionally requires reaching the goal.

For a descriptive binary rate $\hat p=x/n$, we report the 95\% Wilson score
interval
\[
\frac{\hat p+z^2/(2n)\;\pm\;z\sqrt{\hat p(1-\hat p)/n+z^2/(4n^2)}}
     {1+z^2/n},\quad z=1.96.
\]
The appendix tables (Tables \ref{tab:paired-core-significance}, \ref{tab:paired-ablation-significance} and \ref{tab:self-trained-exploratory-significance}) use these Wilson 95\% intervals, and contain more results.  Correction-selection strategies produced success rates of 57\%, 56\%, 46\%, and 51\% for first, all, last, and random selection, respectively. A paired Cochran's $Q$ test found no significant overall difference among the strategies ($p=0.412$); therefore, we performed no post-hoc pairwise comparisons.

\begin{table*}[t]
\centering
\small
\setlength{\tabcolsep}{3pt}
\begin{tabular}{p{0.22\textwidth}c p{0.14\textwidth} p{0.09\textwidth} p{0.14\textwidth} r p{0.20\textwidth}}
\toprule
Domain and representation & Metric & $m{=}0$ rate [95\% CI] & Later checkpoint & Later rate [95\% CI] & $\Delta$ (pp) & Test / $p_{\mathrm{Holm}}$ \\
\midrule
8$\times$8 Grid, without ASCII grid & V & 19\% [12.5, 27.8] & $m{=}60$ & 73\% [65.7, 82.5] & +54 & \(3.18\!\times\!10^{-14}\) \\
8$\times$8 Grid, without ASCII grid & S & 12\% [7.0, 19.8] & $m{=}60$ & 63\% [53.2, 71.8] & +51 & \(9.77\!\times\!10^{-15}\) \\
10$\times$10 Maze, without ASCII grid & V & 7\% [3.4, 13.7] & $m{=}60$ & 57\% [47.2, 66.3] & +50 & \(1.95\!\times\!10^{-14}\) \\
10$\times$10 Maze, with ASCII grid & V & 20\% [13.3, 28.9] & $m{=}60$ & 40\% [30.9, 49.8] & +20 & 0.00164 \\
Sokoban Grid, without ASCII grid & V & 10\% [5.5, 17.4] & $m{=}60$ & 57\% [47.2, 66.3] & +47 & \(9.43\!\times\!10^{-12}\) \\
Full Sokoban, without ASCII grid & V & 12\% [7.0, 19.8] & $m{=}60$ & 42\% [32.8, 51.8] & +30 & \(1.14\!\times\!10^{-6}\) \\
8$\times$8 Grid, with ASCII grid & V & 40\% [30.9, 49.8] & $m{=}60$ & 89\% [81.4, 93.7] & +49 & \(2.86\!\times\!10^{-12}\) \\
8$\times$8 Grid, with ASCII grid & S & 33\% [24.6, 42.7] & $m{=}60$ & 89\% [81.4, 93.7] &+56 & \(3.18\!\times\!10^{-14}\) \\
\bottomrule
\end{tabular}
\caption{Primary/core V/S results. Brackets are marginal Wilson 95\% confidence intervals and are not tests of paired differences. We use the exact two-sided paired McNemar test on complete matched problem IDs, with Holm correction. The paired sample is $n{=}100$.}
\label{tab:paired-core-significance}
\end{table*}

\begin{table*}[t]
\centering
\small
\setlength{\tabcolsep}{2.5pt}
\begin{tabular}{p{0.26\textwidth} p{0.10\textwidth}c p{0.14\textwidth} p{0.14\textwidth} r r}
\toprule
Comparison (A vs.\ B) & Setting & Metric & A rate [95\% CI] & B rate [95\% CI] & $\Delta_{A-B}$ (pp) & $p_{\mathrm{Holm}}$ \\
\midrule
Clean NL corrections vs.\ no corrections & without grid & S & 57\% [47.2, 66.3] & 0\% [0.0, 3.7] & +57 & \(1.39\!\times\!10^{-16}\) \\
Clean NL corrections vs.\ no corrections & with grid & S & 60\% [50.2, 69.1] & 24\% [16.7, 33.2] & +36 & \(3.97\!\times\!10^{-7}\) \\
Clean symbolic feedback vs.\ 50\% corruption & with grid & S & 55\% [45.2, 64.4] & 35\% [26.4, 44.7] & +20 & 0.00497 \\
First-error localization vs.\ random correct example & without grid & S & 30\% [21.9, 39.6] & 8\% [4.1, 15.0] & +22 & \(6.29\!\times\!10^{-5}\) \\
Frozen DeepSeek-V3 prompt: Sonnet vs.\ DeepSeek evaluation & with grid & S & 87\% [79.0, 92.2] & 21\% [14.2, 30.0] & +66 & \(1.40\!\times\!10^{-18}\) \\
Sonnet-4.6 correction-output oracle vs.\ clean symbolic oracle & with grid & S & 31\% [22.8, 40.6] & 55\% [45.2, 64.4] & -24 & \(2.72\!\times\!10^{-4}\) \\
\bottomrule
\end{tabular}
\caption{Correction, corruption, representation, and oracle ablations. $S$ is the percentage of plans that contain no inapplicable action and reach the goal. Marginal intervals are Wilson 95\% intervals.}
\label{tab:paired-ablation-significance}
\end{table*}

\begin{table*}[t]
\centering
\small
\setlength{\tabcolsep}{5pt}
\begin{tabular}{p{0.25\textwidth} p{0.22\textwidth} p{0.22\textwidth} r r}
\toprule
Self-trained model & $m{=}0$ $S$ [95\% CI] & $m{=}60$ $S$ [95\% CI] & $\Delta_{60-0}$ (pp) & $p_{\mathrm{Holm}}$ \\
\midrule
DeepSeek V3 & 6\% [2.8, 12.5] & 27\% [19.3, 36.4] & +21 & \(1.72\!\times\!10^{-5}\) \\
DeepSeek V3.1 & 2\% [0.6, 7.0] & 30\% [21.9, 39.6] & +28 & \(4.93\!\times\!10^{-7}\) \\
Claude Haiku 4.5 & 1\% [0.2, 5.4] & 23\% [15.8, 32.2] & +22 & \(1.43\!\times\!10^{-6}\) \\
Claude Sonnet 4.5 & 10\% [5.5, 17.4] & 61\% [51.2, 70.0] & +51 & \(2.66\!\times\!10^{-15}\) \\
\bottomrule
\end{tabular}
\caption{Exploratory nominal $m{=}0$ versus $m{=}60$ self-trained success comparisons. Every row is without an ASCII grid. Within each model, the same nominally aligned $n{=}100$ problems are compared with exact paired McNemar tests; Holm adjustment is within the full within-model/representation/metric learning-curve family, and brackets are marginal Wilson 95\% intervals.}
\label{tab:self-trained-exploratory-significance}
\end{table*}

\section{Data Appendix}
\label{app:data}

\paragraph{Scope.}
Our evaluation uses five symbolic planning domains: an \(8{\times}8\)
two-room Gridworld, a \(10{\times}10\) Maze, Sokoban Grid without pushable
boxes, Full Sokoban with one movable box, and five-block BlocksWorld.  The
out-of-distribution (OOD) study additionally evaluates a \(15{\times}15\)
Maze using a checkpoint learned on the \(10{\times}10\) Maze.  These
small symbolic domains are useful here because every proposed action can be
checked against an explicit transition system.  Table~\ref{tab:data-pools}
identifies the frozen pools retained for release.  The \(10{\times}10\)
Maze instances share the fixed wall layout in \path{maze.pddl} and differ in
their start and goal cells.

\begin{table*}[t]
\centering
\small
\setlength{\tabcolsep}{2.5pt}
\begin{tabular}{@{}p{0.13\textwidth}p{0.35\textwidth}rrrp{0.20\textwidth}@{}}
\toprule
Task & Canonical data and domain files & Train & Held-out & Eval.\\
\midrule
\(8{\times}8\) Grid
& \path{train/GW_8x8_250_STRIPS.json};
  \path{test/gridworld_250_train_8x8_STRIPS_test.json};
  \path{domain/gridworld-8x8.pddl}
& 250 & 100 & 100  \\
\(10{\times}10\) Maze
& \path{train/pddl_grid.json};
  \path{test/pddl_grid.json};
  \path{domain/maze.pddl}
& 250 & 107 & 100  \\
Sokoban grid (no boxes)
& \path{train/gridworld_sokoban_DZ_final.json};
  \path{test/gridworld_sokoban_DZ_final.json};
  \path{domain/gridworld_sokoban_DZ_final.pddl}
& 250 & 107 & 100  \\
Full Sokoban
& \path{train/sokoban_final.json};
  \path{test/sokoban_final.json};
  \path{domain/sokoban_final.pddl}
& 250 & 127 & 100 \\
BlocksWorld
& \path{train/pddl_UBW.json};
  \path{test/pddl_UBW.json};
  \path{domain/BW_3ops.pddl}
& 250 & 100 & 100  \\
\(15{\times}15\) OOD
& \path{train/15x15_GW_sokoban_1g2w_10x10Cutout.json};
  \path{test/15x15_GW_sokoban_1g2w_10x10Cutout.json};
  \path{domain/gridworld-15x15-sokoban-with-deadzones.pddl}
& 250 & 117 & 100  \\
\bottomrule
\end{tabular}
\caption{Frozen data pools. ``Held-out'' is the number of records in the
stored JSON; ``Eval.'' is the number evaluated.  The \(15{\times}15\) train
pool is archived for completeness but is not the source of the OOD
checkpoint, which was trained on \(10{\times}10\) instances. }
\label{tab:data-pools}
\end{table*}

\paragraph{Splits and identifiers.}
Whenever a held-out JSON contains more than 100 records, we evaluate the
first 100 records in stored-file order. Although each frozen train pool contains
250 records,
reported learning runs consume ordered prefixes of at most 240. In addition, for each task 3 examples were provided in the PTP prompt - these are supplied separately. 
\paragraph{Record and evaluation schema.}
Every task JSON contains metadata and an ordered \texttt{examples} array.
An example has \texttt{input}, containing the complete PDDL problem string,
and \texttt{target}.  The target is intentionally empty in both training
and held-out files: no reference plan is exposed to the model.  Instead, the
symbolic simulator/planner recomputes applicable actions, state
transitions, goal satisfaction, and shortest-path cost.  The unit of
analysis is one problem instance.  Validity \(V\), success \(S\), and
optimality \(O\) are binary per-instance outcomes: a valid plan contains no
inapplicable action, a successful plan is valid and reaches the goal, and
an optimal plan is successful with the minimum number of steps.  Text-only
and ASCII-grid conditions are two renderings of the same symbolic instance,
not different splits.

\paragraph{Generation and preprocessing.}
The following commands show the released generation interfaces. Grid/Maze problem generation samples reachable start--goal
pairs and removes duplicate ordered pairs within one call.  The Sokoban
generator exposes duplicate, triviality, solvability, and plan-length
checks.  The BlocksWorld wrapper invokes \texttt{bwstates} and a PDDL
converter. Further details about the interfaces are found in the codebase accompanying this article.
\begin{lstlisting}[basicstyle=\footnotesize\ttfamily,breaklines=true,columns=fullflexible]
# Grid/Maze interface
python Pddl/grid_utilities/gridworld_toolkit.py \
  gen-problems domain.pddl N hard -o raw_grid/

# BlocksWorld interface (five blocks)
(cd Pddl/pddl-generators/blocksworld && \
 ./generate_blocksworld.sh -n N -b 5 -o raw_bw)

# Sokoban interface; FD enables solvability filtering
python Pddl/sokoban_utils/sokoban_generator.py \
  gen-problem domain.pddl --num-problems N \
  --fd-path "$FD" -o raw_sok/problem.pddl
\end{lstlisting}

For a fixed directory of named raw PDDL files, the common conversion below
sorts filenames, shuffles them with the supplied seed, writes the first 250
to \path{train/TASK.json}, and writes the remainder to
\path{test/TASK.json}.  Seed 42 was used during shuffling - however, as in some cases the seed for generation has been lost, as such the datasets to reconstruct the experiments are released.

\begin{lstlisting}[basicstyle=\footnotesize\ttfamily,breaklines=true,columns=fullflexible]
python Pddl/create_pddl_json.py raw_pddl/ \
  --domain domain.pddl --num_train 250 \
  --task_name TASK --output_dir build/data \
  --seed 42
\end{lstlisting}

The BlocksWorld generator is attributed to the PDDL Generators collection \cite{seipp-et-al-zenodo2022} and its underlying random
state generator to John Slaney and Sylvie Thi\'ebaux \cite{slaney2001blocks}. The gridworld and sokoban problem generators are generated via elementary methods, with the generators and datasets used in the experiments present in the release accompanying this paper. The generators for sokoban and gridworld problems commonly available provide the specification of the grid upfront in the problem.pddl file that is fed into the prompt construction pipeline - to avoid this and abstract away this data through the use of specialised axioms, as well as maintain simplicity in the data processing and simplify domain.pddl generation, the authors opted to implement the generators as opposed to reuse publicly available versions.

  \section{Code Appendix}
  \label{app:code}

  The complete executable code appendix is supplied as the source archive accompanying this paper. It contains all source code, configurations, datasets, and so on. The archive will be released with the paper under a license permitting free research use.
  Appendices~\ref{app:baselines} and~\ref{app:subroutines} exhaustively specify the baseline and implementation procedures.

  The following excerpts show the central localization, correction
  accumulation, and evaluation logic.

  \begin{lstlisting}[language=Python]
  # Localize the first invalid action.
  for idx, action in enumerate(result.predicted_actions):
      applicable = frozenset(
          get_applicable_actions_proxy(state))
      if action not in applicable:
          result.first_invalid_idx = idx
          result.corrections.append(Correction(
              "get_applicable_actions", (state,), applicable))
          break
      state = apply_action_proxy(state, action, goal)

  # Accumulate one rendered example per subroutine/input.
  for result in batch:
      for correction in result.corrections:
          store.setdefault(correction.function, {}) \
               .setdefault(correction.inputs, []) \
               .append(correction.output)

  sample_inputs = list(store[function].keys())
  doctests = generate_doctests(proxy, sample_inputs)

  # Compute the three reported outcomes.
  valid = result.first_invalid_idx is None
  success = valid and result.goal_reached_proxy
  optimal = success and (
      result.predicted_cost == result.optimal_cost)
  \end{lstlisting}



\end{document}